\let\kernellabel\label
\documentclass[aip,cha,12pt]{revtex4-1}
\let\label\kernellabel
\usepackage{graphicx}
\usepackage{microtype}
\usepackage{amssymb}
\usepackage{amsmath}
\usepackage{xcolor}
\usepackage{dcolumn}% Align table columns on decimal point
\definecolor{mygray}{gray}{0.4}
\usepackage{threeparttable}
\usepackage{array}
\usepackage{paralist}
\usepackage{multirow}
\usepackage{enumerate} % Advanced list customization
\usepackage{hyperref}
\hypersetup{
    colorlinks=true,
    linkcolor=blue,
    filecolor=magenta,      
    urlcolor=cyan,
    pdfcreator={},
    pdfproducer={}
}

\usepackage{enumitem}
\usepackage{adjustbox}
\usepackage{newfloat}
\DeclareFloatingEnvironment[name=Algorithm,placement=htbp]{algorithm}
\usepackage{algpseudocode}
\begin{document}

%\preprint{}

\title{Stochastic Engrams for Efficient Continual Learning} %Title of paper


\author{Isabelle Aguilar}
\email{isabelle.aguilar@sydney.edu.au}

\author{Luis Fernando Herbozo Contreras}
\email{luis.herbozocontreras@sydney.edu.au}

\author{Omid Kavehei}
\email{omid.kavehei@sydney.edu.au}

\affiliation{School of Biomedical Engineering, The University of Sydney, Camperdown, NSW, Australia, 2006}

% \author{}
% %\email[]{Your e-mail address}
% %\homepage[]{Your web page}
% %\thanks{}
% %\altaffiliation{}
% \affiliation{}


\date{\today}

\begin{abstract}
The ability to learn continuously in artificial neural networks (ANNs) is often limited by catastrophic forgetting, a phenomenon in which new knowledge becomes dominant. By taking mechanisms of memory encoding in neuroscience (i.e., engrams) as inspiration, we propose a novel approach that integrates stochastically-activated engrams as a gating mechanism for metaplastic binarized neural networks ({\it m}BNNs). This method leverages the {\it computational efficiency of {\it m}BNNs} combined with {\it the robustness of probabilistic memory traces} to mitigate forgetting and maintain the model's reliability. Previously validated metaplastic optimization techniques have been incorporated to further enhance synaptic stability. Compared to baseline binarized models and benchmark fully connected continual learning approaches, our method is the only strategy capable of achieving average accuracies over 70\% in both class-incremental and domain-incremental MNIST benchmarks, matching full-precision state-of-the-art methods. Furthermore, we achieve a significant reduction in peak GPU and RAM usage, under 5\% and 20\%, respectively, as well as an $\sim 8\times$ reduction in memory footprint compared to full precision counterparts. Our findings demonstrate {\bf (A)} an improved stability {\it vs.} plasticity trade-off, {\bf (B)} reduced memory intensiveness, and {\bf (C)} enhanced performance in binarized architectures. By uniting principles of neuroscience and efficient computing, we offer new insights into the design of scalable and robust deep learning systems.

\end{abstract}

\pacs{}% insert suggested PACS numbers in braces on next line

\maketitle %\maketitle must follow title, authors, abstract and \pacs

%% Main Body %%
\section{Introduction}
\label{intro}
Machine learning (ML) has made remarkable advances in solving complex problems across various domains, from natural language processing \cite{lauriola2022introduction} to autonomous systems \cite{bachute2021autonomous, shinde2018review}. These successes have been primarily driven by the training and fine-tuning of neural networks on large, static, and retrospective datasets. However, in dynamic and prospective environments where data arrives sequentially, often in time-sensitive, real-time applications, these networks face a fundamental limitation known as \textit{catastrophic forgetting} \cite{mccloskey1989catastrophic}. 

Artificial neural networks (ANNs) tend to lose previously acquired knowledge when training on new tasks. They often cannot adapt to incoming data, making them unsuitable for applications that require continuous learning, such as implantable medical devices with closed-loop feedback, where strong energy constraints, wireless data telemetry bandwidth, and onboard memory capacity are crucial \cite{lee2019overview, salam2016tradeoffs}. These applications also exist in other domains where local data processing is optional, for example, on edge devices with adaptive response needs or closed-loop feedback, or are necessary, for instance, on autonomous unmanned underwater vehicles for civil and defense purposes, where any signal emission or communication is either undesired or impossible \cite{mcgann2008adaptive}. In most cases, these operations also involve long-term periods in which the device encounters previously unseen environments, underscoring the importance of continual learning \cite{wang2024comprehensive, lee2020clinical}. 
Such implementations require the intentional use of their limited resources. With this perspective, it is instrumental to examine more unconventional architectures that promote efficiency. Here, binarized neural networks (BNNs) pose a unique advantage due to their computational efficiency compared to that of the full-precision multilayer perceptron (MLP), although they remain uncommon backbones for continual learning \cite{courbariaux2015binaryconnect}.

On the other hand, biological neural systems possess an extraordinary ability to learn continuously by retaining previously acquired memories while seamlessly integrating new information \cite{o2012computational}. The key mechanism underlying this capability is the observation that the brain uses sparse, context-modulated neural substrates, called engrams, for memory consolidation and retrieval \cite{rao2021synaptic, guskjolen2023engram, autore2023adaptive}. The selective gating of these engrams, driven by contextual signals \cite{josselyn2015finding}, regulates neural population activity during learning and recall, providing a natural blueprint for lifelong learning.

Despite the clear parallel, existing continual learning methods rely heavily on full-precision models that are computationally prohibitive for resource-constrained edge applications. Additionally, while BNNs offer extreme energy efficiency, their rigid, discrete parameter space has left them fundamentally underutilized and underperforming in sequential learning scenarios. 

\begin{figure*}[ht!]
\centering
\includegraphics[width=0.90\textwidth]{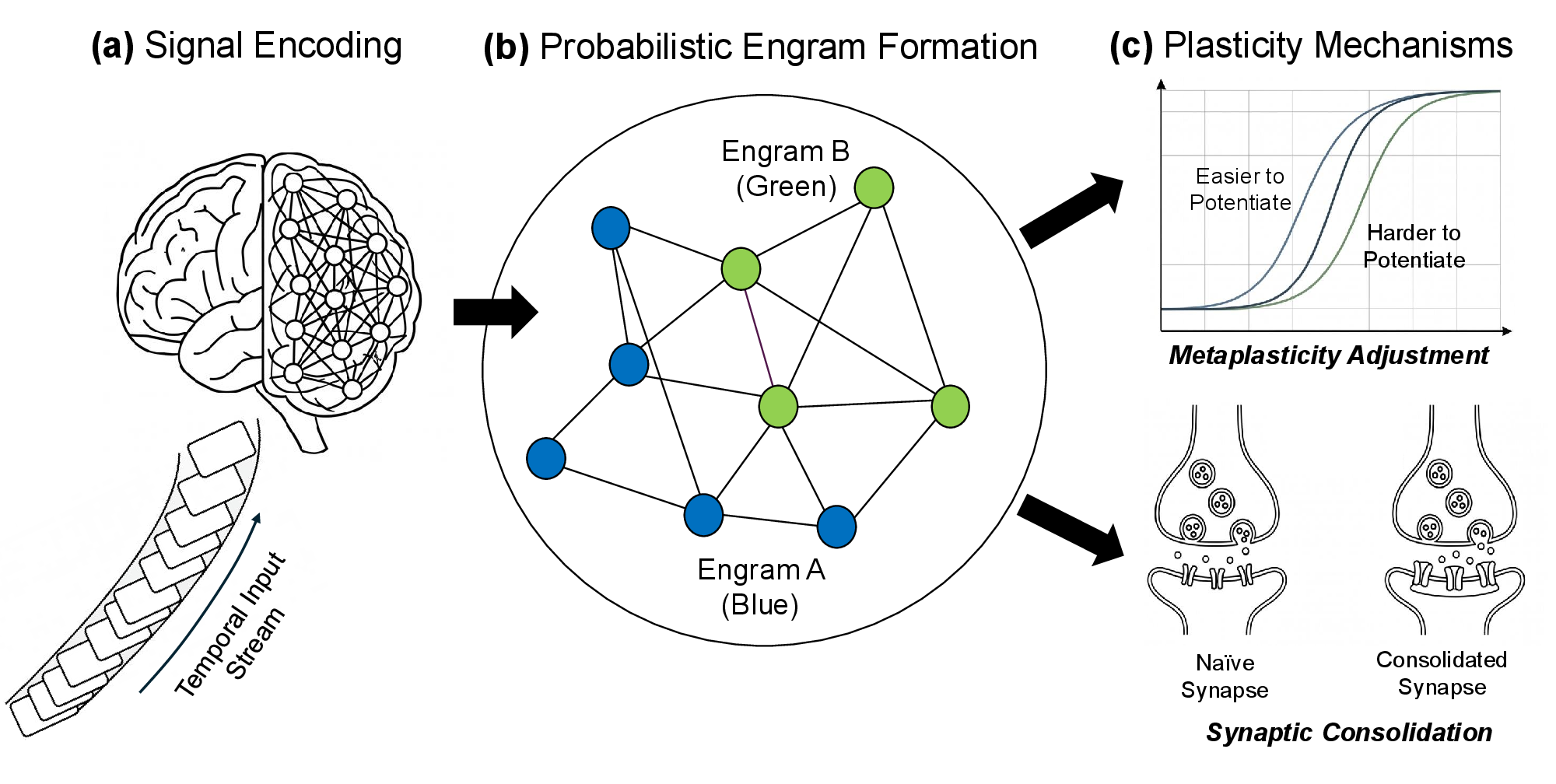}
\caption{Continual learning mechanisms in biological and artificial networks. \textbf{(a)} In the brain, varying experiences and stimuli are streamed into the system over time, \textbf{(b)} forming different engrams (green and blue) that encode and consolidate unique information. \textbf{(c)} As incoming memories flood the network, metaplasticity and synaptic plasticity allow the brain to adapt to additional knowledge. This process repeats throughout life, and our brains continually learn.}
\label{fig:fig1} 
\end{figure*}

To the best of our knowledge, \textit{this work presents the first framework to integrate context-dependent engram gating directly within a latent space for a metaplastic binarized architecture.} Our core novelty lies in the formulation of context-dependent, engram-inspired gating that operates directly on a binarized latent space, unlocking robust memory retention without sacrificing computational advantages. Fig.~\ref{fig:fig1} illustrates the biological motivations behind the model, which we denote as \textit{engramBNN}.

The performance of our method is evaluated in class-incremental and domain-incremental settings and compared with both baseline and relevant state-of-the-art methods \cite{masse2018alleviating, kirkpatrick2017overcoming, zenke2017continual, li2017learning, laborieux2021synaptic, shi2025hybrid, bricken2023sparse, kang2022forget}. For class-incremental experiments, we use the Split MNIST benchmark with a 5-2 split (5 tasks, each with 2 new, unseen classes). For a more realistic and difficult task, we also perform Split CIFAR-100 at a 20-5 and 10-10 split (detailed in Sec.~\ref{sec:sub:class}). For domain-incremental experiments, we employ the realistic CORe50 dataset \cite{lomonaco2017core50} and the Split MNIST benchmark (detailed in Sec.~\ref{sec:sub:domain}).
The class- and domain-incremental simulations use a single shared classifier across tasks (unlike the multi-head setup used in task-incremental learning). Consequently, task identity is not implicitly provided via separate output heads. For baselines that explicitly rely on task context or task-specific sub-networks such as XdG \cite{masse2018alleviating} (which conditions hidden units via task-dependent gating vectors) and WSN \cite{kang2022forget} (which selects its task-specific binary mask using a task index during inference), the task identity is provided to activate the corresponding gate/mask. In contrast, for methods evaluated under strict task-agnostic settings, task identity remains completely unknown at evaluation, ensuring a rigorous benchmark against catastrophic forgetting \cite{farquhar2018towards}. 

In summary, the main contributions of this work are:
\begin{itemize}
    \item \textbf{A Bio-Inspired BNN Architecture:} We propose \textit{engramBNN}, an energy-efficient architecture that combines binarized networks with neuro-inspired metaplasticity and dynamic engram gating, enhancing memory retention and reducing memory overhead.
    \item \textbf{Latent-Space Engram Gating:} We introduce a simple gating block that leverages encoded signals to gate active neurons in sequential learning scenarios and prevents catastrophic overwriting of valuable weights by using features extracted from a compressed latent space.
    \item \textbf{State-of-the-Art Resource Efficient Performance:} We demonstrate the effectiveness of the proposed approach on various continual learning benchmarks, showing competitive performance with state-of-the-art methods and outperforming the \textit{m}BNN baseline accuracy by $\sim$ 10\%, 1.5\%, and 3\% in Domain-IL Split MNIST, Domain-IL CORe50, and both 10-10 and 20-5 Class-IL Split CIFAR-100, respectively, while maintaining $\sim 8\times$ less memory cost than existing methods.
\end{itemize}

\section{Related Works}
\label{relatedworks}

Continual learning methods typically fall into five categories below \cite{wang2024comprehensive}, which can potentially overlap with each other:
\begin{inparaenum}[(1)]
    \item regularization-based approaches \cite{kirkpatrick2017overcoming, zenke2017continual, li2017learning},
    \item rehearsal or replay-based approaches \cite{van2020brain, rebuffi2017icarl, lopez2017gradient},
    \item architectural-based approaches \cite{mallya2018piggyback, mallya2018packnet},
    \item optimization-based approaches \cite{zeng2019continual, javed2019meta}, and
    \item representation-based approaches \cite{javed2019meta, madaan2021representational}.
\end{inparaenum}

\subsection{Neurobiological Processes}

Metaplasticity, or the plasticity of synaptic plasticity, is a neurobiological plasticity modulator, also described briefly as the "plasticity of synaptic plasticity" \cite{abraham1996metaplasticity, abraham2008metaplasticity}. This biological mechanism regulates plasticity by influencing synaptic states and the brain's adaptive capacity to incoming information \cite{fusi2005cascade}. This ability is essential for balancing the varying strengths of synapses, stabilizing incoming knowledge in the brain while allowing for and adapting to new memories. This process is an attractive regularization or optimization approach for ANNs and has proven successful for continual learning compared to baseline feedforward networks \cite{jedlicka2022contributions}.

Another key mechanism in the brain is the formation and recall of engrams, which are sparse memory traces that encode, consolidate, and retrieve specific memories \cite{guskjolen2023engram}. These neural ensembles are sparsely distributed across the brain and play a crucial role in recalling past information. Rodent-based experiments have observed the natural reactivation of these specific engrams during memory retrieval \cite{reijmers2007localization, tayler2013reactivation}. Furthermore, direct intervention studies using optogenetics have demonstrated that artificially activating these tagged neural ensembles is sufficient to causally cue memory retrieval \cite{liu2012optogenetic}.

Engrams are also closely related to associative learning, a fundamental area of study in neuroscience and psychology, including the notable case of Pavlov's dog in classical conditioning \cite{wasserman1997whats}. In neuroscience, associative learning interprets how stimulus representations in the brain (formed by cues, environmental contexts, temporal information, etc.) inform behavioral outcomes \cite{thompson1997associative}. As memories and information continue to stream into the brain, associated contexts help form engrams, update our biological neural networks, and adapt to incoming representations \cite{heald2023computational}. Furthermore, the brain has long been viewed as a stochastic system, and more recent studies have even proposed that this stochasticity enhances associative memories and retention \cite{gershman2015unifying, liljenstrom1995noise}.

\subsection{Bio-inspired Continual Learning Strategies}

These biological processes have inspired numerous methodologies in deep learning to address the discrepancy between artificial and biological networks in continual learning \cite{jeon2023distinctive, hadsell2020embracing, jedlicka2022contributions}. Several existing deep learning strategies share conceptual analogies with engram formation, although they are not explicitly modelled. For instance, the isolated subnetworks within an ensemble approach serve as a functional parallel to separate engrams \cite{dong2020survey}. Similarly, dropout and sparse neural networks exhibit analogous behaviors where specific groups of active neurons act as loosely inspired memory traces for a given task \cite{louizos2017learning, srivastava2014dropout}. Transfer learning and recurrent neural networks (RNNs) also employ associative memory mechanisms that improve storage compared to traditional multilayer perceptrons (MLPs) \cite{pan2009survey, rizzuto2001autoassociative}. Beyond structural analogies, stochasticity has been explored via probabilistic activations \cite{tang2013learning} and stochastic synapses sampled from Bernoulli distributions, which have demonstrated competitive continual learning performance while also addressing energy efficiency \cite{schug2021presynaptic}. 

Our method builds on this concept of stochastic modularity but extends it into a context-dependent framework. Rather than applying stochastic activations directly to the model's weights, our stochasticity lies in the distinct gating mechanism applied to the backbone network. In terms of engram-based computational research, the use of context-dependent gating to allocate neuronal ensembles represents the most direct translation of associative learning into an architecture-based continual learning approach \cite{masse2018alleviating, tilley2023artificial}. While previous biologically inspired models have explored computational efficiency, the specific integration of explicit engram gating mechanisms within extremely quantized binarized neural networks (BNNs) for severe resource-constrained environments remains unexplored. Other biologically inspired continual learning models typically focus on different strategy types, such as replay-based or distinct architecture-based models.

\subsection{Related Architectures}

In the ML field, binarized neural networks (BNNs) have emerged as a competitive alternative to their full-precision counterparts \cite{courbariaux2015binaryconnect,conti2018xnor, hubara2016binarized}. These one-bit networks show promise in producing a low-power solution to the catastrophic forgetting problem \cite{laborieux2021synaptic}, and are particularly attractive due to their hardware compatibility, although most of these works rely on replay-based mechanisms that impose more memory constraints \cite{basso2024class, vorabbi2024enabling}. Recent methods have continued to implement metaplasticity models in efficient architectures \cite{zohora2021metaplasticnet, cho2025neuromimetic}. Our model leverages the metaplasticity function first introduced in \cite{laborieux2021synaptic} but differs from these methods by using a single classifier (rather than the ensemble network employed in \cite{zohora2021metaplasticnet}) and adding an engram module to incorporate context-dependent gating within the network.
Similarly, autoencoders and variational autoencoders (VAEs) have been previously employed for generative continual learning using latent spaces, such as brain-inspired replay \cite{ye2023continual, deja2021binplay, van2020brain}. However, these strategies are mainly replay-based, imposing significant memory demands that scale with increasing data volume. Additionally, modular architectures similar to biological neurogenesis have also been explored for continual learning, but they are difficult to scale as memory demands increase with the number of tasks \cite{rusu2016progressive}. We propose an alternative method that addresses the memory constraints these approaches overlook, and primarily utilizes an encoder module.

\section{Methods}
\label{methods}

\subsection{Binarized Neural Network}
\label{sec:sub:bnn}
The backbone of our model is an efficient BNN rather than a typical fully connected, full-precision network. These layers are shown as yellow blocks within the overarching architecture in Fig.~\ref{fig:fig2}. BNNs are highly quantized, bitwise versions of standard MLPs that dramatically reduce memory and energy usage while achieving performance comparable to or even better than full-precision models \cite{simons2019review, rastegari2016xnor}. Unlike typical feedforward networks, the activations and weights of a BNN are bounded to $\{-1$,$+1\}$. Backpropagation operates on a real-valued weight to influence gradient updates. However, the evaluation stage uses the binary weight, which equals the sign of the real-valued weight (i.e., the "hidden weight").

\subsection{Engram Gating Block}
\label{sec:sub:engram} 

\begin{figure*}[ht!]
\centering
\includegraphics[width=1.0\textwidth]{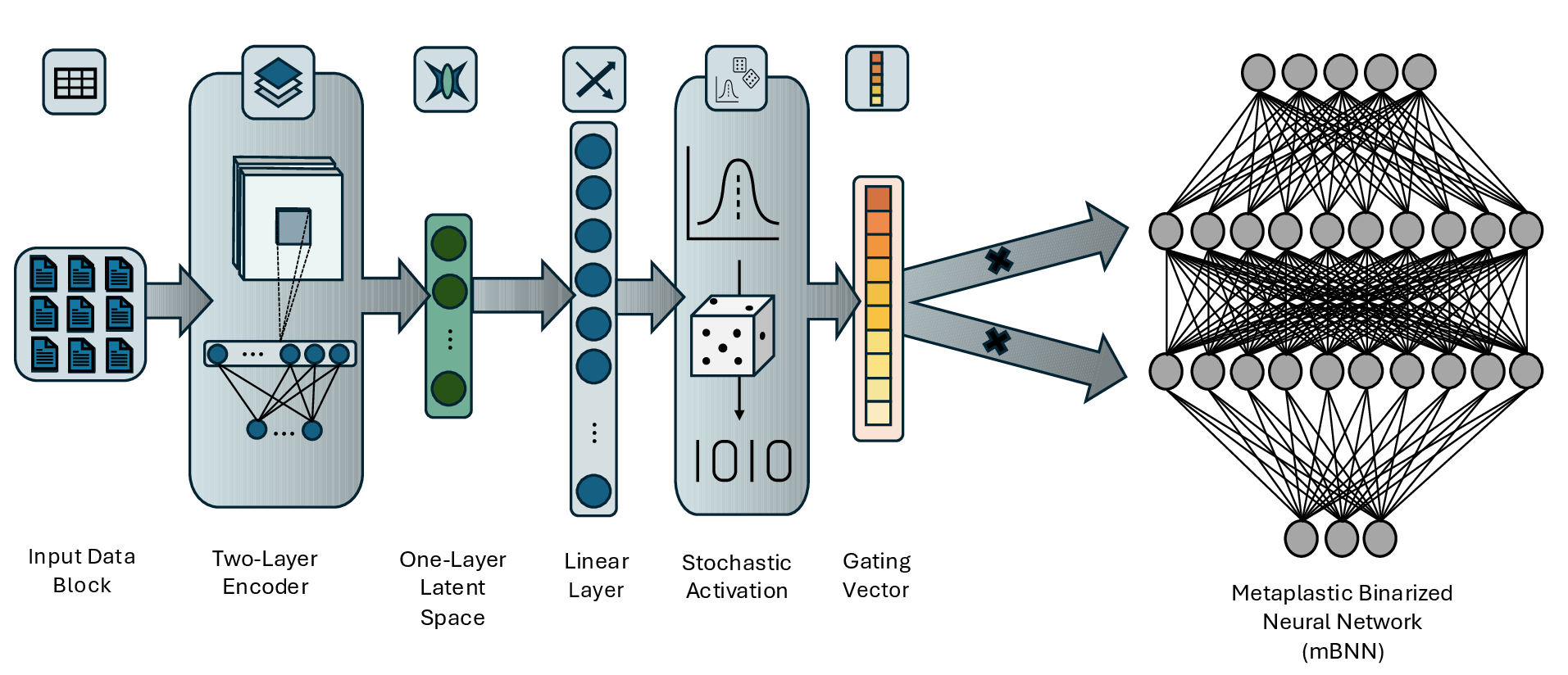}
\caption{\textbf{Engram block architecture.} The engram block processes the input data block through a two-layer encoder to compress features into a one-layer latent space. This compressed representation is then passed through a linear layer and subjected to stochastic activation, yielding a discrete gating vector. Finally, this gating vector modulates the mBNN via element-wise multiplication across its hidden layers, dynamically routing the forward pass.}
\label{fig:fig2} 
\end{figure*}

To promote task-specific representations and mitigate interference during continual learning, we introduce a stochastic engram-gating block that modulates the activity of hidden neurons via context-dependent gating. This component is illustrated in blue and expanded upon in Fig.~\ref{fig:fig2}.

For a given input sample, a single shared two-layer full-precision encoder projects it into a lower-dimensional latent space to extract essential spatial context. From this shared latent space, separate linear layers project outward to generate distinct, independent gating vectors (masks) for each corresponding hidden layer of the metaplastic BNN (\textit{m}BNN). Because the gating is context-dependent, a unique mask is dynamically generated for every individual input on every forward pass.

While the internal encoder utilizes ReLU activations, the final projection layers instead apply a stochastic activation and a top-$k$ mask to generate discrete {$0,1$} gating values. Both the encoder and linear layers are fully trainable, and their parameters are updated jointly with the rest of the network. While the architecture bears a structural resemblance to an autoencoder by distilling meaningful representations of the input, its functional objective diverges; the embedding drives task-specific modulation rather than input reconstruction.

\textbf{Stochastic Routing and Top-\textit{k} Sparsity.}
To generate the discrete binary mask, a stochastic activation initiates the engrams that subsequently modulate the hidden neurons. We first compute the continuous activation probabilities $p$ using a sigmoid function applied to the pre-activation logit $z$ (Equ.~\ref{eq:1}). To strictly enforce sparsity and limit the capacity used per task, we apply a top-$k$ operation directly to these probabilities. The parameter $k$ determines the exact sparsity level and is defined as a fixed hyperparameter across all tasks for a given dataset, chosen after initial validation (evaluating retention ranges from 10\% to 30\%). Specifically, for the gating layer, $k$ is set to retain 20\% of the units (e.g., $k=409$ for a 2048-dimensional layer), which corresponds directly to the 80\% sparsity constraint reported in our subsequent evaluations.

As defined in Equ.~\ref{eq:2}, the top-$k$ operation creates a sparse probability vector $\tilde{p}$, masking out all but the $k$ highest probabilities of $p$. For each unit $i$:

\begin{eqnarray}
    p & =\sigma(z) = \frac{1}{1 + e^{-z}}  \label{eq:1} \\
    \tilde{p_i} & = \begin{cases} 
      p_i & \text{if } p_i \text{ is in top-}k \text{ of } p \\
      0 & \text{otherwise}
   \end{cases} \label{eq:2} \\
    {M} & \sim \text{Bernoulli}(\tilde{p}) 
    \label{eq:3}
\end{eqnarray}

During training, Bernoulli sampling is applied element-wise to $\tilde{p}$ to yield a discrete binary mask $M \in \{0, +1\}$ (Equ.~\ref{eq:3}), while all non-top-$k$ units are strictly zeroed out. While the routing mechanism activates at most $k$ units per input, the active count remains tightly bounded near $k$ as the top probabilities rapidly saturate toward $1.0$. Additionally, for large $k$ (e.g., $k=409$), the risk of complete signal dropout is astronomically small ($0.5^k$ at initialization), ensuring stable forward propagation.

\textbf{Inference and Gradient Propagation.}
During inference, we bypass stochastic Bernoulli sampling to ensure deterministic and stable behavior. Instead, we apply a hard deterministic top-$k$ selection, setting the units with the highest gating logits directly to $+1$ and the rest to $0$. Since the fully binarized network trains exclusively on discrete $\{0, 1\}$ signals, active units consistently output $+1$, eliminating the need for probability scaling. We use deterministic inference to efficiently evaluate the exact engram assigned to a given input.

The resulting vector $M$ is multiplied element-wise by the activations of each \textit{m}BNN hidden layer. During the backward pass in training, gradients are routed only through the active top-$k$ units and propagated through the stochastic activation using a straight-through estimator \cite{hubara2016binarized}.

Unlike Dropout \cite{srivastava2014dropout}, which is a regularization method that applies random masking during training, our engram block remains active during both training and testing to ensure consistent task-specific routing. This mechanism encourages dynamic sparsity within the backbone architecture and modulates activity across tasks, thus reducing overlapping representations and catastrophic interference. Furthermore, Dropout does not use context to guide its masking, whereas the engram block employs embeddings to apply context-dependent gating to the backbone.

\subsection{Metaplasticity Function}  \label{sec:sub:metaplasticity} 
Artificial metaplasticity is implemented as a second continual learning component within the BNN backbone. Implemented during the optimization process, this mechanism emulates synaptic plasticity and metaplasticity, thereby mitigating catastrophic forgetting. Metaplasticity in ANNs has been proven to be a powerful asset to continual learning \cite{aguilar2025continuous, laborieux2021synaptic}. A function of metaplasticity, introduced by Laborieux {\it et al.} \cite{laborieux2021synaptic}, is defined in Equ.~\ref{eq:fmeta}. The hyperparameter $m$ controls the network's tendency to consolidate weights towards a task,  which in turn affects its sensitivity to new data. $W_h$ is the real-valued hidden weight, whose absolute value signifies synaptic importance. In other words, as $|W_h|$ increases, it becomes more difficult to flip the binary sign of the synapse.

\begin{equation}
f_{\rm meta}(m, W_h) = 1 - \tanh^2(m \cdot W_h)
\label{eq:fmeta}
\end{equation}

A detailed breakdown of the metaplastic weight updates is outlined in Algorithm~\ref{alg:metaplastupdate}. Here, the standard weight update, $U_W$, is applied to the binarized weights. This update informs the metaplastic modulation of $W_h$ via $f_{meta}$ based on the model's current synaptic state.

As it is strengthened and tuned, the hyperparameter $m$ presents a stability-plasticity trade-off. Demonstrated in \cite{aguilar2025continuous, laborieux2021synaptic}, a higher $m$ value helps strengthen the retention of older tasks, but the network fails to reach earlier peak performance on newer tasks as a result.While large $m$ values eventually drive $f_{\rm meta}$ near zero, this numerical saturation is intended to freeze consolidated knowledge without prematurely restricting learning. Given our initialization width of 0.01 (expected $\vert{}W_h\vert{} \approx 0.005$), the initial modulation factor remains large ($f_{\rm meta} \in [0.28, 0.50]$) for robust early plasticity. As $\vert{}W_h\vert{}$ grows, the function naturally bounds updates to safeguard prior engrams.

%% Algorithm 1
    % Define a custom color named "mygrey"
    \definecolor{mygray}{gray}{0.4}
    \begin{algorithm}[htbp]
    \caption{Hidden weight metaplastic update for BNN, described in our previous work in \cite{aguilar2025continuous}. $\mathbf{W_h}$ is the vector of real-valued hidden weights, and $W_h$ denotes one element (as applied to all other vectors involved), $\mathbf{U_W}$ is the weight update produced by the Adam gradient descent algorithm, ($x$,$y$) is a batch of labeled training data, m is the metaplasticity parameter, and $\eta$ is the learning rate.}
    \label{alg:metaplastupdate}
    \begin{algorithmic}[1]  % The [1] argument gives line numbers

    \Require $\mathbf{W_h}$, ($x$, $y$), $\mathbf{U_W}$, $\eta$, $m$
    
    \State $\mathbf{W_b}\leftarrow\text{Sign}(\mathbf{W_h})$  
    \Comment\textcolor{mygray}{Computing binary weights}
    
    \State $\hat{y}\leftarrow\text{Forward}(x,\mathbf{W_b})$  
    \Comment\textcolor{mygray}{Perform inference}
    
    \State $C\leftarrow\text{Cost}(\hat{y},y)$  
    \Comment\textcolor{mygray}{Compute mean loss over the batch}
    
    \State $\mathbf{U_W}\leftarrow\text{Adam Backward}(C,\mathbf{W_b}, \mathbf{U_W},\hat{y})$ 
    \Comment\textcolor{mygray}{Compute backward update with the binarized weights}
    
    \If{$U_W\cdot W_b>0$}  
    \Comment\textcolor{mygray}{For a single weight if $U_W$ prescribes to decrease $|W_h|$}
        \State $W_h\leftarrow W_h-\eta\mathbf{U_W}\cdot f_{\text{meta}}(m,W_h)$  
        \Comment\textcolor{mygray}{Metaplastic update}
    \Else
        \State $W_h\leftarrow W_h -\eta\mathbf{U_W}$
    \EndIf
        \State
        \Return $W_h$, $U_W$
    
    \end{algorithmic}
    \end{algorithm}

\section{Experiment}
\label{sec:exp}

\subsection{Implementation Details}

Our model utilizes a binarized backbone network. For standard experiments, this consists of two hidden layers, each containing 2048 neurons, with engram gating projected onto both linear layers. For Split CIFAR-100 experiments, a binarized ResNet-18 is employed. Here, engram gating is spatially broadcast to modulate the convolutional feature maps. More specifically, it is applied after the first batch normalization and again at the output of each residual block. 
Experience replay \cite{rolnick2019experience} was implemented using a fixed buffer of $2000$ samples. To ensure a fair and rigorous comparison, every CIFAR baseline receives this exact same replay buffer capacity and sampling budget. The buffer is updated via standard reservoir sampling, and during training, replay samples are mixed with current task data at a 1:1 ratio within a standard batch size of $128$, following standard preprocessing (outlined in Appendix~\ref{secA2}). All the methods we compared used the same backbone architecture, except for the state-of-the-art SDMLP, CH-HNN, and WSN, whose architectures are integral to their strategies. 
In the Split MNIST setting, the metaplastic hyperparameter $m$ is set to 5; in Split CIFAR-100, it is set to 250. For CORe50, $m$ is set equal to 175. For more stable training, batch-normalization is added after each layer \cite{courbariaux2015binaryconnect}. The real-valued hidden weights are initialized over a uniform distribution between the values $[-0.5,0.5]$. The learning rate was set to $\eta = 10^{-4}$ during training. Adam's beta values $(\beta_1,\beta_2)$ are $(0.9,0.999)$, respectively, as default.

In the engram block architecture, the computation graph is distinctly divided into a shared representation encoder and a gating projection network. For our standard MLP backbone, the shared encoder processes the input through a sequence of fully connected layers with decreasing dimensionalities ($\text{input} \to 512 \to 128 \to 64$). This bottleneck design enables the compression of arbitrary data types and facilitates robust representation learning in the latent space. Subsequently, the gating projection maps this condensed latent state to the exact width of the backbone layers ($64 \to 2048 \to 2048$) to generate the final masking probabilities. 

The model is trained for 20 epochs per task for all incremental settings. For the CORe50 experiment, we increase the capacity of the backbone by widening the hidden layers from 2048 to 4096 neurons. Consequently, the mask width is scaled equivalently; the last engram projection layer is adjusted at 4096 neurons to match the wider backbone. Furthermore, to prevent cross-task statistical interference during continual learning, batch normalization statistics (i.e., running mean and variance) are isolated: they are explicitly saved at the end of each task's training phase and reloaded during task-specific inference. The performance of the models is evaluated on the test set throughout training.

Our model, termed engramBNN, is tested and compared with a vanilla BNN baseline. Continual learning baselines are also compared with our model and applied to the BNN architecture. We examine the performance of BNNs with context-dependent gating (XdG) \cite{masse2018alleviating}, elastic weight consolidation (EWC) \cite{kirkpatrick2017overcoming}, synaptic intelligence (SI) \cite{zenke2017continual}, and learning without forgetting (LwF) \cite{li2017learning}. 

Additionally, recent biologically inspired state-of-the-art methods are tested, a hybrid neural network based on cortico-hippocampal recurrent loops (CH-HNN) \cite{shi2025hybrid} and a sparsely distributed memory multilayer perceptron based on the cerebellum (SDMLP) \cite{bricken2023sparse}. Winning Sub-Networks (WSN) is also considered a state-of-the-art method that learns and selects subnetworks within its model for each task \cite{kang2022forget}. The CH-HNN model is excluded from domain-incremental learning experiments because it requires pre-trained ANN models and CORe50 embeddings that were not available in the original experiments.

Serving as an ablation study comparison, we compare the performance, sparsity, and stability of the engramBNN, metaplastic BNN ({\it m}BNN) \cite{laborieux2021synaptic} without the gating module attached, and a BNN with Dropout regularization. This provides a fair comparison of the engram block's benefits in our network. All analyses are conducted with the class-incremental Split MNIST experimental setting.

In this paper, we compare our model with conceptually similar approaches that focus on regularization and gating mechanisms to facilitate continual learning. However, our model efficiency comparison is done with a broader range of strategies. We compare our model with both replay and modular strategies, specifically brain-inspired replay (BI Replay) \cite{van2020brain} and progressive neural networks (PNN) \cite{rusu2016progressive}, as well as a network with stochastic synapses \cite{schug2021presynaptic} and a full-precision version of the engramBNN.

\subsection{Metrics}

In lifelong learning settings, earlier performance metrics are further specified in the cumulative and final point versions. Let $A_{j,k}$ denote the model's test accuracy on task $k$ after having sequentially trained up to task $j$. Average accuracy (ACC) is reported. As shown in Equ.~\ref{eq:acc}, ACC takes the final accuracy, $A_{T,i}$, for each task $i$ and averages them over the number of tasks seen. This gives a task-wise view of learning retention of the continual learning network.

\begin{equation}
{\rm ACC}=\frac{1}{T}\sum_{i=1}^{T} A_{T,i}
\label{eq:acc}
\end{equation}

A weighted accuracy value ({\it w}ACC) is also calculated. The standard ACC metric treats all tasks equally, which can obscure catastrophic forgetting if a model performs well on only the most recently learned tasks. To address this, \textit{w}ACC prioritizes long-term memory stability and acts as a stress test rather than a balanced metric. It explicitly isolates and heavily penalizes the forgetting of the earliest tasks. As defined in Equ.~\ref{eq:w_acc}, the final accuracy for each task $i$ after training on $T$ total tasks (denoted as $A_{T,i}$) is scaled by a weight $w_i$, and then normalized over the sum of all weights.

The weighting factor is defined by an exponential function, $w_i = e^{\lambda t_i}$, where $t_i$ represents the "age" of the task relative to the end of training (calculated as $T - i$) with a decay rate hyperparameter set to $\lambda = 0.5$. Because of this exponential scaling, the metric induces a deliberately steep gradient (e.g., in a 20-task sequence, the oldest task dominates the final score while the most recent tasks contribute marginally). In conjunction with ACC, \textit{w}ACC acts as a lower-bound indicator of lifelong retention, ensuring that a high score cannot be achieved by merely excelling on the newest data at the expense of the oldest.

\begin{equation}
{\rm {\it w}ACC}=\frac{\sum_{i=1}^{T} w_i A_{T,i}}{\sum_{i=1}^{T} w_i}
\label{eq:w_acc}
\end{equation}

Forward Transfer (FWT) and Backward Transfer (BWT) are also continual learning metrics that evaluate learning and forgetting, respectively \cite{lopez2017gradient}. FWT examines forward knowledge transfer and evaluates a model's ability to generalize learning from past tasks to future ones. It is defined as:

\begin{equation}
{\rm FWT}=\frac{1}{T-1} \sum_{i=2}^{T} A_{i-1,i} 
\label{eq:fwt}
\end{equation}

\noindent
where $A_{i-1,i}$ is the accuracy on task $i$ evaluated immediately after the network has finished training on task $i-1$ (before any exposure to the training data of task $i$).

BWT quantifies how prior task knowledge is retained or forgotten during subsequent task learning. It is defined as:

\begin{equation}
{\rm BWT}=\frac{1}{T-1} \sum_{i=1}^{T-1} \left( A_{T,i} - A_{i,i} \right)~.
\label{eq:bwt}
\end{equation}

A positive BWT signifies improved knowledge when learning new tasks, while a negative BWT signifies knowledge deterioration. As it is standard behavior for models to show catastrophic forgetting, we also add a modified metric based on BWT, known as Remembering (REM). Remembering or ${\rm REM}=1-|\min({\rm BWT},0)|$.

\subsection{Domain-Incremental Learning Experiments} \label{sec:sub:domain}

% CORe50 & Split MNIST Table
\begin{table*}[t]
    \caption{Domain-incremental experiment metrics on the CORe50 and MNIST datasets. Average values are reported in percentages along with their respective standard deviations.}
    \label{tab:tab1}
    \footnotesize % Reduces overall font size to help it fit
    \setlength{\tabcolsep}{3pt} % Reduces horizontal padding between columns
    \begin{ruledtabular}
    \begin{tabular}{llccccc}
        \textbf{Model} & \textbf{Experiment} & \textbf{ACC (\%) $\uparrow$} & \textbf{\textit{w}ACC$^\ast$ (\%) $\uparrow$} & \textbf{FWT (\%) $\uparrow$} & \textbf{BWT (\%) $\uparrow$} & \textbf{REM (\%) $\uparrow$} \\
        \hline
        Vanilla BNN & CORe50-NI & $\mathbf{20.94} \pm {\scriptstyle 0.23}$ & $11.94 \pm {\scriptstyle 0.42}$ & $14.33 \pm {\scriptstyle 0.37}$ & $-82.43 \pm {\scriptstyle 02.52}$ & $17.57 \pm {\scriptstyle 02.52}$\\
         & Split MNIST & $61.41 \pm {\scriptstyle 0.39}$ & $58.71 \pm {\scriptstyle 0.40}$ & $53.77 \pm {\scriptstyle 0.93}$ & $-42.87 \pm {\scriptstyle 28.67}$ & $57.13 \pm {\scriptstyle 28.67}$ \\
        \hline
        \textbf{engramBNN} & CORe50-NI & $17.88 \pm {\scriptstyle 0.13}$ & $\mathbf{19.06} \pm {\scriptstyle 0.21}$ & $\mathbf{16.49} \pm {\scriptstyle 0.37}$ & $-17.40 \pm {\scriptstyle 15.98}$ & $82.60 \pm {\scriptstyle 15.98}$ \\
         & Split MNIST & $\mathbf{75.34} \pm {\scriptstyle 1.83}$ & $\mathbf{79.47} \pm {\scriptstyle 2.61}$ & $52.20 \pm {\scriptstyle 1.03}$ & $-25.58 \pm {\scriptstyle 18.06}$ & $74.42 \pm {\scriptstyle 18.06}$ \\
        \hline
        EWC & CORe50-NI & $19.82 \pm {\scriptstyle 0.46}$ & $11.43 \pm {\scriptstyle 0.59}$ & $12.63 \pm {\scriptstyle 0.59}$ & $\mathbf{2.23} \pm {\scriptstyle 1.11}$ & $\mathbf{100.00} \pm {\scriptstyle 1.11}$ \\
         & Split MNIST & $66.19 \pm {\scriptstyle 0.98}$ & $64.12 \pm {\scriptstyle 1.29}$ & $53.25 \pm {\scriptstyle 2.40}$ & $-41.11 \pm {\scriptstyle 27.02}$ & $58.89 \pm {\scriptstyle 27.02}$\\
        \hline
        SI & CORe50-NI & $10.51 \pm {\scriptstyle 0.27}$ & $11.22 \pm {\scriptstyle 0.27}$ & $12.34 \pm {\scriptstyle 0.43}$ & $0.75 \pm {\scriptstyle 0.83}$ & $100.00 \pm {\scriptstyle 0.83}$ \\
         & Split MNIST & $50.47 \pm {\scriptstyle 4.09}$ & $50.14 \pm {\scriptstyle 3.58}$ & $49.25 \pm {\scriptstyle 10.10}$ & $\mathbf{-1.44} \pm {\scriptstyle 1.52}$ & $\mathbf{98.56} \pm {\scriptstyle 1.52}$\\
        \hline
        LwF & CORe50-NI & $19.17 \pm {\scriptstyle 00.15}$ & $11.08 \pm {\scriptstyle 00.18}$ & $12.27 \pm {\scriptstyle 0.47}$ & $-82.46 \pm {\scriptstyle 2.67}$ & $17.54 \pm {\scriptstyle 2.67}$ \\
         & Split MNIST & $60.79 \pm {\scriptstyle 0.35}$ & $54.28 \pm {\scriptstyle 0.36}$ & $53.76 \pm {\scriptstyle 0.67}$ & $-47.36 \pm {\scriptstyle 28.01}$ & $52.64 \pm {\scriptstyle 28.01}$\\
        \hline
        XdG & CORe50-NI & $17.88 \pm {\scriptstyle 0.09}$ & $10.13 \pm {\scriptstyle 0.09}$ & $11.25 \pm {\scriptstyle 0.43}$ & $-55.95 \pm {\scriptstyle 5.19}$ & $44.05 \pm {\scriptstyle 5.19}$ \\
         & Split MNIST & $68.19 \pm {\scriptstyle 0.50}$ & $64.71 \pm {\scriptstyle 0.69}$ & $54.59 \pm {\scriptstyle 0.88}$ & $-35.78 \pm {\scriptstyle 22.42}$ & $64.22 \pm {\scriptstyle 22.42}$ \\
        \hline
        \textit{m}BNN & CORe50-NI & $16.24 \pm {\scriptstyle 0.30}$ & $14.88 \pm {\scriptstyle 0.44}$ & $13.47 \pm {\scriptstyle 0.49}$ & $-53.38 \pm {\scriptstyle 19.60}$ & $46.62 \pm {\scriptstyle 19.60}$ \\
         & Split MNIST & $65.08 \pm {\scriptstyle 0.87}$ & $59.47 \pm {\scriptstyle 0.74}$ & $54.13 \pm {\scriptstyle 1.43}$ & $-42.47 \pm {\scriptstyle 26.74}$ & $57.53 \pm {\scriptstyle 26.74}$ \\
        \hline
        SDMLP & CORe50-NI & $18.30 \pm {\scriptstyle 0.63}$ & $18.13 \pm {\scriptstyle 0.50}$ & $13.78 \pm {\scriptstyle 0.86}$ & $-3.70 \pm {\scriptstyle 2.74}$ &  $96.30 \pm {\scriptstyle 2.74}$ \\
         & Split MNIST & $69.58 \pm {\scriptstyle 0.79}$  & $68.31 \pm {\scriptstyle 1.31}$ & $\mathbf{55.58} \pm {\scriptstyle 0.74}$ & $-30.32 \pm {\scriptstyle 20.04}$ &  $69.68 \pm {\scriptstyle 20.04}$ \\
        \hline
        WSN & CORe50-NI & $10.74 \pm {\scriptstyle 0.34}$ & $10.72 \pm {\scriptstyle 0.35}$ & $12.18 \pm {\scriptstyle 0.93}$ & $-9.34 \pm {\scriptstyle 23.37}$ &  $90.66 \pm {\scriptstyle 23.37}$ \\
         & Split MNIST & $51.92 \pm {\scriptstyle 4.02}$ & $50.51 \pm {\scriptstyle 4.69}$ & $49.39 \pm {\scriptstyle 7.05}$ & $-10.32 \pm {\scriptstyle 18.15}$ & $89.68 \pm {\scriptstyle 18.15}$\\
    \end{tabular}
    \end{ruledtabular}
    \vspace{1ex}
    \raggedright
    {\scriptsize \textsuperscript{a} For each model, all performance results are averaged over six training runs. The best values are in bold. $^\ast$\textbf{\textit{w}ACC}: weighted ACC.}
\end{table*}

Continual learning encompasses various types of incremental learning: domain-incremental, class-incremental, and task-incremental. Each addresses different challenges in adapting to new data distributions. In domain-incremental learning, the input distribution shifts over time while the task and output targets remain unchanged. This differs from class-incremental learning, where new categories are introduced sequentially, and task-incremental learning, where new tasks require distinct boundaries. Here, we evaluate our model across class- and domain-incremental settings to assess the robustness of our model to various dataset biases and environmental shifts. 

The most notable domain-incremental benchmark task is the Permuted MNIST task \cite{zenke2017continual}. In this scenario, the model must learn various permutations of the MNIST image dataset sequentially. Each task has a unique and random spatial permutation of pixels. Here, we argue that although widely used in the continual learning field, Permuted MNIST has fundamental flaws that render it an unreliable indicator of continual learning ability. There have been previous criticisms of Permuted MNIST, as it does not translate to real-world scenarios and is, at best, a toy problem that does not capture the true nature of realistic domain shifts \cite{farquhar2018towards}. We present Permuted MNIST results in Sec.~\ref{secA1} to examine our model under a simple domain-incremental task.

Instead, we test our model on the more realistic CORe50-NI tasks ("NI" meaning "New Instances"), a specific experimental setting of the CORe50 dataset, where each new instance is imposed by a domain shift defined by varying backgrounds, lighting, poses, and occlusions \cite{lomonaco2017core50}. The CORe50-NI dataset was captured across 11 sessions with unique backgrounds and lighting conditions. For our configuration, the network targets the 10 broader category classes (aggregating the 50 underlying objects). The model is trained sequentially across eight sessions (treated as individual continuous learning tasks) and tested on the remaining three. Unlike task-incremental settings where each task $k$ possesses a distinct test split, these three held-out sessions form a single, unified test set. Consequently, the performance metric after learning task $j$ (analogous to $A_{j,k}$) is obtained by evaluating the model on the entirety of this fixed test set, yielding a global measure of domain generalization and retention. This dataset, in particular, has become more recognized as a benchmark in domain-incremental learning in recent years and has been curated with continual learning in mind \cite{shi2023unified}. As a supplementary experiment, we perform the domain-incremental version of Split MNIST, in which the model must classify digits $0$ through $9$ as odd or even across all tasks, which each contain two of the ten total classes.

As shown in Table~\ref{tab:tab1}, in the CORe50-NI experiment, although vanilla BNN achieved the highest recorded ACC, our engramBNN model achieved the highest weighted ACC of $19.06\%$. Our model also had the highest FWT score of $16.49\%$. Interestingly, the baselines EWC and SI achieved the best BWT and REM scores, being the only models with positive BWT scores. In the Split MNIST domain-incremental experiments, our model achieved the highest ACC and weighted ACC of $75.34\%$ and $79.47\%$, respectively. SI exhibited the best BWT and REM scores, and SDMLP had the highest FWT score. It should be noted that BWT exhibits higher variance than ACC across the six runs because the discrete nature of BNNs can cause sudden, localized task collapses that heavily penalize BWT, but are diluted in the overall ACC average.

\begin{figure*}[ht!]
\centering
\includegraphics[width=1.0\textwidth]{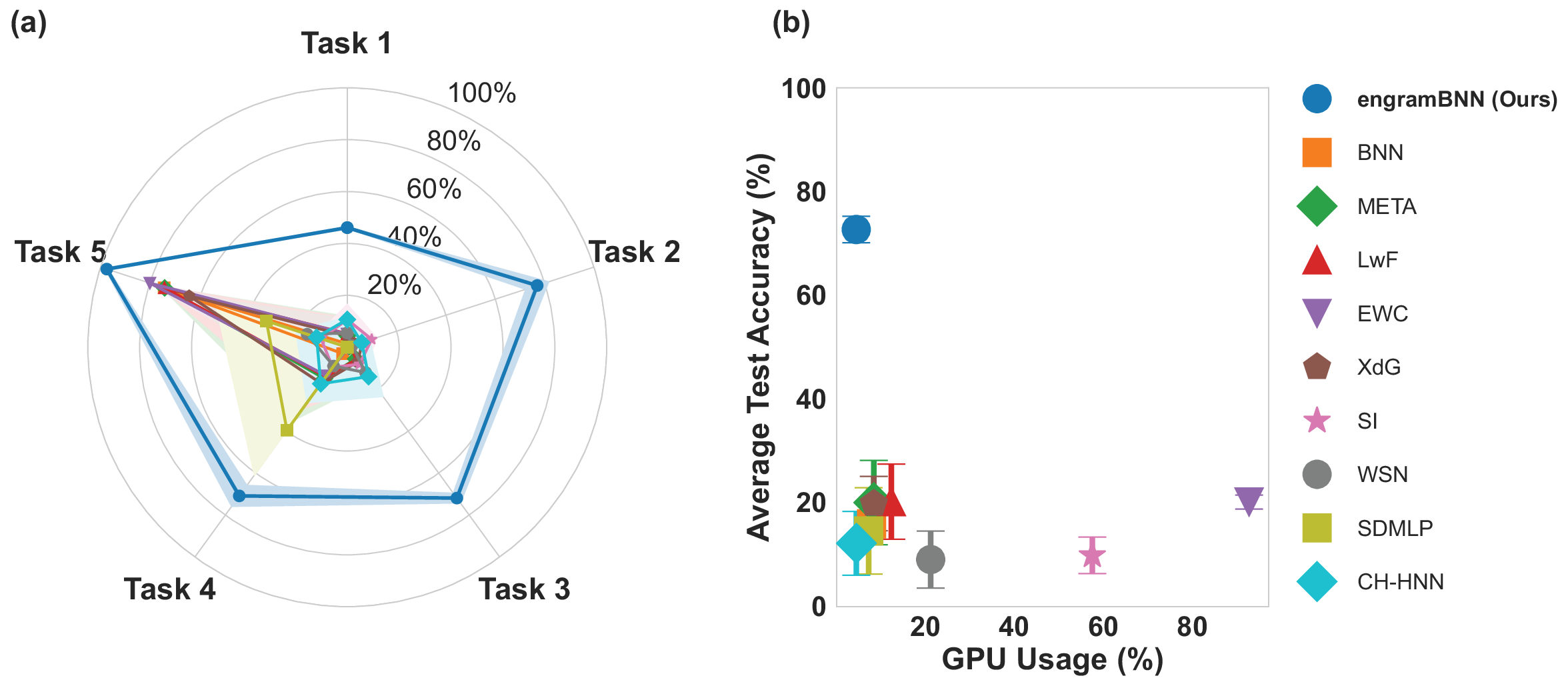}
\caption{\textbf{Performance and resource efficiency comparison of engramBNN against baseline CL models.} On the Class-IL Split MNIST benchmark, we present a \textbf{(a)} radar chart illustrating the final test accuracy on each of the five individual tasks after sequentially learning the entire task sequence, highlighting retention across early tasks. \textbf{(b)} Trade-off between final overall average test accuracy and computational overhead, quantified by peak GPU memory utilization in percent Error bars denote standard deviation.}
\label{fig:fig3} 
\end{figure*}

\subsection{Class-Incremental Experiments}
\label{sec:sub:class}

Another test for lifelong learning is class-incremental learning. Here, classes are introduced incrementally. The Split MNIST experiment divides the standard MNIST dataset into a sequence of tasks, each containing a distinct subset of classes. Following the typical baseline setup, our test consists of five tasks, each introducing two new classes that have not been previously seen. Similarly, Split CIFAR divides the CIFAR-100 dataset. Compared to MNIST, which consists of grayscale images of handwritten digits across one of 10 classes, CIFAR-100 is much more complex and contains RGB images of varied natural objects, such as animals, vehicles, and plants, that fall under one of 100 classes. Two experiments are conducted on this dataset: a 20-5 split (20 tasks with 5 unseen classes per task) and a 10-10 split. Similarly to domain-incremental experiments, although distinct tasks are set up, task IDs are not provided to the model, meaning our model is unaware when these tasks shift from one to the other.

% Split MNIST Table
\begin{table*}[t]
    \caption{Class-incremental Split MNIST continual learning metrics. Average results are reported along with their standard deviations.}
    \label{tab:tab2}
    \begin{ruledtabular}
    \begin{tabular}{lccccc}
        \textbf{Model} & \textbf{ACC (\%) $\uparrow$} & \textbf{\textit{w}ACC$^\ast$ (\%) $\uparrow$} & \textbf{FWT (\%) $\uparrow$} & \textbf{BWT (\%) $\uparrow$} & \textbf{REM (\%) $\uparrow$}
        \\
        \hline
        Vanilla BNN & $15.83 \pm {\scriptstyle 0.32}$ & $8.18 \pm {\scriptstyle 0.19}$ & $5.23 \pm {\scriptstyle 3.31}$ & $-70.33 \pm {\scriptstyle 3.58}$ & $29.67 \pm {\scriptstyle 3.58}$\\
        \textbf{engramBNN} & $\mathbf{72.65 \pm {\scriptstyle 1.42}}$ & $\mathbf{76.31 \pm {\scriptstyle 2.30}}$ & $0.55 \pm {\scriptstyle 0.13}$ & $-33.10 \pm {\scriptstyle 11.96}$ & $66.90 \pm {\scriptstyle 11.96}$ \\
        EWC & $20.05 \pm {\scriptstyle 0.77}$ & $10.99 \pm {\scriptstyle 0.59}$ & $0.04 \pm {\scriptstyle 1.87}$ & $-99.25 \pm {\scriptstyle 1.04}$ & $0.75 \pm {\scriptstyle 1.04}$\\
        SI & $9.76 \pm {\scriptstyle 1.69}$ & $9.42 \pm {\scriptstyle 1.48}$ & $\mathbf{9.29 \pm {\scriptstyle 2.88}}$ & $\mathbf{-0.22 \pm {\scriptstyle 0.77}}$ & $\mathbf{99.78 \pm {\scriptstyle 0.77}}$\\
        LwF & $20.12 \pm {\scriptstyle 3.97}$ & $12.03 \pm {\scriptstyle 4.06}$ & $4.08 \pm {\scriptstyle 4.48}$ & $-64.77 \pm {\scriptstyle 3.96}$ & $35.23 \pm {\scriptstyle 3.96}$\\
        XdG & $19.74 \pm {\scriptstyle 2.45}$ & $13.97 \pm {\scriptstyle 2.57}$ & $4.64 \pm {\scriptstyle 3.97}$ & $-52.38 \pm {\scriptstyle 3.94}$ & $47.62 \pm {\scriptstyle 3.94}$ \\
        \textit{m}BNN & $29.94 \pm {\scriptstyle 4.71}$ & $11.84 \pm {\scriptstyle 4.39}$ & $4.55 \pm {\scriptstyle 6.16}$ & $-53.74 \pm {\scriptstyle 4.67}$ & $46.26 \pm {\scriptstyle 4.67}$ \\
        SDMLP & $14.47 \pm {\scriptstyle 5.90}$  & $9.85 \pm {\scriptstyle 4.08}$ & $0.00 \pm {\scriptstyle 0.17}$ & $-52.93 \pm {\scriptstyle 36.24}$ &  $47.07 \pm {\scriptstyle 36.24}$ \\
        CH-HNN & $12.08 \pm {\scriptstyle 3.17}$ & $10.71 \pm {\scriptstyle 3.43}$ & $4.51 \pm {\scriptstyle 8.93}$ & $-11.71 \pm {\scriptstyle 9.87}$ & $88.29 \pm {\scriptstyle 9.87}$ \\
        WSN & $8.94 \pm {\scriptstyle 2.89}$ & $7.33 \pm {\scriptstyle 3.29}$ & $8.53 \pm {\scriptstyle 2.45}$ & $-23.61 \pm {\scriptstyle 40.87}$ & $76.39 \pm {\scriptstyle 40.87}$\\
    \end{tabular}
    \end{ruledtabular}
    \vspace{1ex}
    \raggedright
    {\scriptsize \textsuperscript{a} For each model, all performance results are averaged over six training runs. The best values are in bold. $^\ast$\textbf{\textit{w}ACC}: weighted ACC.}
\end{table*}

% Split CIFAR Table
\begin{table*}[t]
    \caption{Class-incremental Split CIFAR-100 continual learning metrics. Average results are reported along with their standard deviations.}
    \label{tab:tab3}
    \footnotesize % Reduces overall font size to help it fit
    \setlength{\tabcolsep}{1.3pt} % Reduces the blank space between columns
    \begin{ruledtabular}
    \begin{tabular}{lcccccc}
        \textbf{Model} & \textbf{Task-Class Split} & \textbf{ACC (\%) $\uparrow$} & \textbf{\textit{w}ACC$^\ast$ (\%) $\uparrow$} & \textbf{FWT (\%) $\uparrow$} & \textbf{BWT (\%) $\uparrow$} & \textbf{REM (\%) $\uparrow$}
        \\
        \hline
        Vanilla BNN & $20-5$ & $4.05 \pm {\scriptstyle 0.25}$ & $0.50 \pm {\scriptstyle 0.39}$ & $0.02 \pm {\scriptstyle 0.47}$ & $-50.13 \pm {\scriptstyle 7.30}$ & $49.87 \pm {\scriptstyle 7.30}$ \\ 
         & $10-10$ & $5.90 \pm {\scriptstyle 0.17}$ & $2.74 \pm {\scriptstyle 0.43}$ & $0.01 \pm {\scriptstyle 0.55}$ & $-47.90 \pm {\scriptstyle 4.40}$ & $52.10 \pm {\scriptstyle 4.40}$ \\ 
        \hline
        \textbf{engramBNN} & $20-5$ & $\mathbf{8.12} \pm {\scriptstyle 0.25}$ & $\mathbf{3.74} \pm {\scriptstyle 0.43}$ & $\mathbf{0.61} \pm {\scriptstyle 0.40}$ & $-16.89 \pm {\scriptstyle 11.24}$ & $83.11 \pm {\scriptstyle 11.24}$ \\
         & $10-10$ & $\mathbf{9.66} \pm {\scriptstyle 0.24}$ & $\mathbf{4.58} \pm {\scriptstyle 0.55}$ & $\mathbf{0.58} \pm {\scriptstyle 0.46}$ & $-16.95 \pm {\scriptstyle 7.05}$ & $83.05 \pm {\scriptstyle 7.05}$ \\ 
        \hline
        EWC & $20-5$ & $3.30 \pm {\scriptstyle 0.10}$ & $0.00 \pm {\scriptstyle 0.01}$ & $0.00 \pm {\scriptstyle 0.38}$ & $-61.60 \pm {\scriptstyle 6.63}$ & $38.40 \pm {\scriptstyle 6.63}$\\ 
         & $10-10$ & $5.39 \pm {\scriptstyle 0.11}$ & $0.00 \pm {\scriptstyle 0.00}$ & $0.00 \pm {\scriptstyle 0.50}$ & $-48.72 \pm {\scriptstyle 4.33}$ & $51.28 \pm {\scriptstyle 4.33}$\\ 
        \hline 
        SI & $20-5$ & $3.73 \pm {\scriptstyle 0.21}$ & $0.59 \pm {\scriptstyle 0.21}$ & $0.03 \pm {\scriptstyle 0.54}$ & $-47.23 \pm {\scriptstyle 6.78}$ & $52.77 \pm {\scriptstyle 6.78}$\\ 
         & $10-10$ & $5.21 \pm {\scriptstyle 0.27}$ & $0.56 \pm {\scriptstyle 0.21}$ & $0.01 \pm {\scriptstyle 0.40}$ & $-42.81 \pm {\scriptstyle 04.91}$ & $57.19 \pm {\scriptstyle 4.91}$\\ 
        \hline
        LwF & $20-5$ & $7.71 \pm {\scriptstyle 0.24}$ & $1.29 \pm {\scriptstyle 0.20}$ & $0.02 \pm {\scriptstyle 0.57}$ & $-47.36 \pm {\scriptstyle 5.84}$ & $52.64 \pm {\scriptstyle 5.84}$\\
         & $10-10$ & $6.07 \pm {\scriptstyle 0.20}$ & $1.29 \pm {\scriptstyle 0.18}$ & $0.02 \pm {\scriptstyle 0.57}$ & $-48.11 \pm {\scriptstyle 4.75}$ & $51.89 \pm {\scriptstyle 4.75}$\\ 
        \hline
        XdG & $20-5$ & $3.98 \pm {\scriptstyle 0.25}$ & $1.27 \pm {\scriptstyle 0.16}$ & $1.02 \pm {\scriptstyle 0.49}$ & $-51.09 \pm {\scriptstyle 6.34}$ & $48.91 \pm {\scriptstyle 6.34}$ \\
         & $10-10$ & $6.03 \pm {\scriptstyle 0.19}$ & $1.27 \pm {\scriptstyle 0.16}$ & $0.02 \pm {\scriptstyle 0.33}$ & $-47.61 \pm {\scriptstyle 4.60}$ & $52.39 \pm {\scriptstyle 4.60}$ \\ 
        \hline 
        \textit{m}BNN & $20-5$ & $5.23 \pm {\scriptstyle 0.30}$ & $0.92 \pm {\scriptstyle 0.16}$ & $0.03 \pm {\scriptstyle 0.71}$ & $-44.73 \pm {\scriptstyle 7.30}$ & $55.27 \pm {\scriptstyle 7.30}$ \\ 
         & $10-10$ & $6.33 \pm {\scriptstyle 0.26}$ & $1.92 \pm {\scriptstyle 0.35}$ & $0.02 \pm {\scriptstyle 0.45}$ & $-43.82 \pm {\scriptstyle 4.32}$ & $56.18 \pm {\scriptstyle 4.32}$ \\
        \hline
        SDMLP & $20-5$ & $1.00 \pm {\scriptstyle 0.00}$  & $0.00 \pm {\scriptstyle 0.00}$ & $0.00 \pm {\scriptstyle 0.00}$ & $\mathbf{0.00} \pm {\scriptstyle 0.00}$ &  $\mathbf{100.00} \pm {\scriptstyle 0.00}$ \\ 
         & $10-10$ & $1.00 \pm {\scriptstyle 0.00}$  & $0.00 \pm {\scriptstyle 0.00}$ & $0.00 \pm {\scriptstyle 0.00}$ & $\mathbf{0.00} \pm {\scriptstyle 0.00}$ &  $\mathbf{100.00} \pm {\scriptstyle 0.00}$ \\
        \hline 
        CH-HNN & $20-5$ & $2.22 \pm {\scriptstyle 0.22}$ & $3.58 \pm {\scriptstyle 0.40}$ & $0.02 \pm {\scriptstyle 3.26}$ & $-3.27 \pm {\scriptstyle 5.53}$ & $96.73 \pm {\scriptstyle 5.53}$ \\ 
         & $10-10$ & $2.14 \pm {\scriptstyle 0.22}$ & $2.48 \pm {\scriptstyle 0.40}$ & $0.01 \pm {\scriptstyle 0.66}$ & $-3.90 \pm {\scriptstyle 5.14}$ & $96.10 \pm {\scriptstyle 5.14}$ \\ 
        \hline
        WSN & $20-5$ & $0.95 \pm {\scriptstyle 0.25}$ & $0.97 \pm {\scriptstyle 0.59}$ & $0.58 \pm {\scriptstyle 0.32}$ & $-3.39 \pm {\scriptstyle 14.53}$ & $96.61 \pm {\scriptstyle 14.53}$\\ 
         & $10-10$ & $1.03 \pm {\scriptstyle 0.22}$ & $0.97 \pm {\scriptstyle 0.59}$ & $0.58 \pm {\scriptstyle 0.35}$ & $-5.90 \pm {\scriptstyle 16.84}$ & $94.10 \pm {\scriptstyle 16.84}$\\ 
    \end{tabular}
    \end{ruledtabular}
    \vspace{1ex}
    \raggedright
    {\scriptsize \textsuperscript{a} For each model, all performance results are averaged over six training runs. The best values are in bold. $^\ast$\textbf{\textit{w}ACC}: weighted ACC.}
\end{table*}

In Table~\ref{tab:tab2}, the engramBNN method had both higher ACC and Weighted ACC at $72.65\%$ and $76.31\%$, respectively. SI demonstrated better retention of prior knowledge, with higher BWT and REM scores and a higher FWT score. 

To evaluate the performance and computational efficiency of engramBNN further, we visualized the performance accuracy and resource efficiency on the Class-IL Split MNIST benchmark in Fig.~\ref{fig:fig3}. As shown in Fig.~\ref{fig:fig3}a, the radar chart illustrates the final test accuracy across all five sequentially learned tasks after the completion of the full training sequence. While baseline methodologies suffer from severe catastrophic forgetting and collapses toward low accuracy levels (typically under $20\text{--}40\%$) across earlier tasks, engramBNN maintains consistently high accuracy across every task. Furthermore, Fig.~\ref{fig:fig3}b highlights the efficiency-accuracy trade-off by plotting the final overall average test accuracy against peak GPU memory utilization (\%). While competing baseline algorithms either achieve low accuracy despite low resource footprint (i.e., CH-HNN, \textit{m}BNN, WSN) or require substantial computational overhead without accuracy gains (i.e., EWC which utilizes upwards of $90\%$ GPU usage and SI exceeding $55\%$), engramBNN occupies the optimal region of high accuracy and low GPU usage. Specifically, our model achieves the highest overall average test accuracy ($\sim73\%$) while maintaining a minimal computational footprint ($\sim5\%$ peak GPU usage), validating engramBNN as a strategy to effectively mitigate catastrophic forgetting without incurring prohibitive hardware overhead.

The results for the Split CIFAR-100 experiments are presented in Table~\ref{tab:tab3}. For this benchmark, Binarized convolutional networks were used, compared to the Binarized multilayer perceptron used in the simpler Split MNIST; Single-headed, class-incremental learning on CIFAR-100 with a small sampling buffer (representing $4\%$ of the entire training set) is a notoriously difficult task. When further constrained by the discrete parameter space of binarized networks, all baseline models naturally struggled to capture and retain the dataset's complexities. 

Despite these severe constraints, engramBNN achieved the highest absolute accuracy (ACC) and weighted accuracy (\textit{w}ACC) across both the $20-5$ and $10-10$ split simulations. Specifically, engramBNN achieved a $100\%$ relative accuracy improvement over the Vanilla BNN in the 20-5 split ($8.12\%$ vs. $4.05\%$) and comfortably outperformed established regularization baselines such as EWC ($3.30\%$), SI ($3.73\%$), and XdG ($3.98\%$). Furthermore, while LwF was most similar to our model's accuracy in the 20-5 split, its performance degraded in the 10-10 sequence ($6.07\%$), whereas engramBNN demonstrated superior scalability by climbing to $9.66\%$ accuracy.

Methods such as SDMLP, CH-HNN, and WSN demonstrate artificially high REM scores (near $100.00\%$) and minimal BWT loss. However, in this context, it does not indicate successful continual learning. Rather, it is symptomatic of total plasticity collapse. Because these models fail to acquire new knowledge beyond the initial task (reflected by an ACC of $0.95-2.22\%$, some of which are equivalent to random chance), there is no newly learned representation to interfere with prior tasks. In contrast, engramBNN provides the strongest available trade-off by successfully acquiring new classes while maintaining the highest BWT and REM among models that actively learn.

\subsection{Ablation and Analysis of the Engram Block}
\label{sec:sub:engramanalysis}

Ablation experiments were conducted on the class-incremental Split MNIST setting. Test accuracy, sparsity, and mask stability are reported in Table~\ref{tab:tab4}. Sparsity is measured during deterministic inference as the fraction of inactive neurons (masked or set to 0) in the second hidden layer of the network when evaluated on the test set. Mask stability was measured for the engramBNN to quantify how well the model preserved task-specific representations. For each task $i$, a binary task-level mask $M_i^{(j)}$ is extracted, providing information on which neurons remain active for that task after learning task $j$. Because engram gating is input-dependent, this task-level mask is defined as the union of all active neurons across the entirety of task $i$'s evaluation set. The reference mask $M_i^{(i)}$ corresponds to this union mask immediately after training on task $i$. Stability is computed as:

\begin{equation}
\text{Stability}(i, j) = \frac{|M_i^{(i)} \cap M_i^{(j)}|}{|M_i^{(i)}|}
\end{equation}

\noindent
where $|\cdot|$ denotes cardinality (i.e., the total number of active neurons) in the given mask or intersection. To strictly measure retention and avoid artificially inflating the average, the baseline comparison at $j=i$ (which guarantees a match of $1.0$) is excluded. Therefore, mask stability is computed exclusively across all subsequent tasks $j > i$. The resulting stability values were averaged within each task, and then averaged across all tasks to yield a singular metric, Average Mask Stability.

\begin{table}[t]
    \caption{\textbf{Ablation Metrics}. All metrics are averaged over 6 runs. Best values are reported in bold.}
    \label{tab:tab4}
    \begin{ruledtabular}
    \begin{tabular}{l c c c}
        \textbf{Model} & \textbf{Accuracy ($\%$) $\uparrow$} & \textbf{Sparsity $\uparrow$} & \shortstack{\textbf{Average Mask} \\ \textbf{Stability} $\uparrow$} \\
        \hline
        \textbf{engramBNN} & $\mathbf{72.06 \pm {\scriptstyle 2.09}}$ & $\mathbf{0.80 \pm {\scriptstyle 0.00}}$ & $\mathbf{0.70 \pm {\scriptstyle 0.00}}$ \\
        \textit{m}BNN & $30.20 \pm {\scriptstyle 2.33}$ & $0.00 \pm {\scriptstyle 0.00}$ & N/A \\
        \begin{tabular}{@{}l@{}}BNN with \\ Dropout\end{tabular} & $20.94 \pm {\scriptstyle 3.88}$ & $0.00 \pm {\scriptstyle 0.00}$ & N/A \\
    \end{tabular}
    \end{ruledtabular}
\end{table}

\begin{figure}[ht!]
\centering
\includegraphics[width=0.95\textwidth]{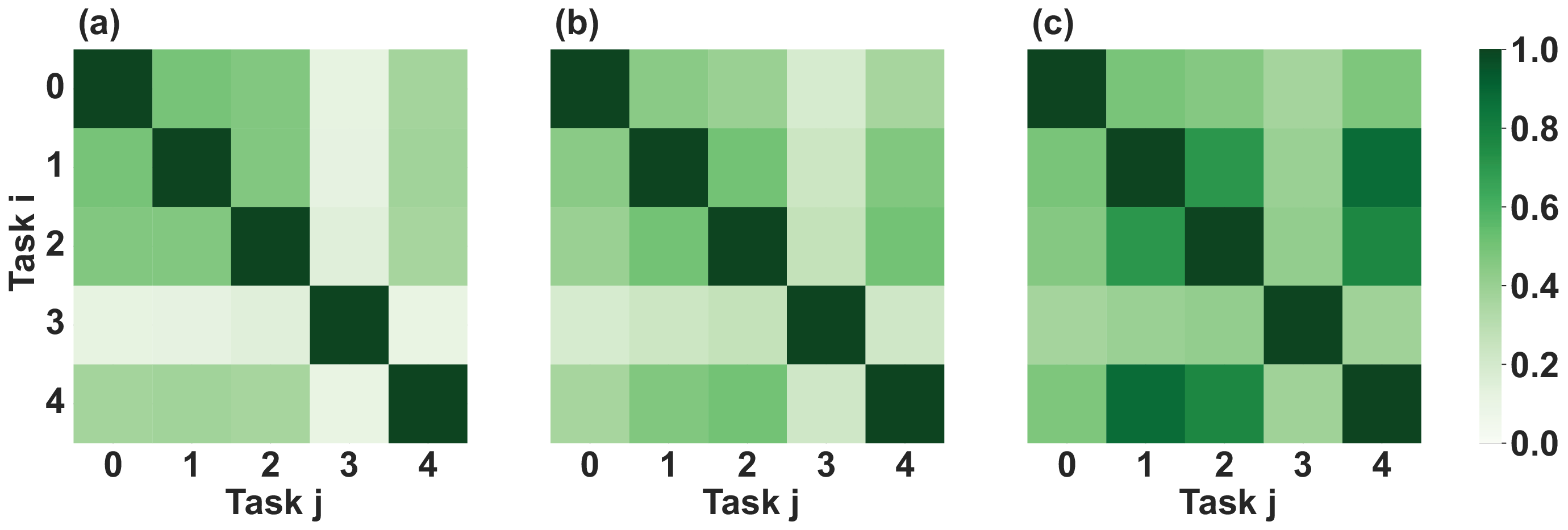}
\caption{\textbf{Task Jaccard Overlap of Neuron Masks}. Jaccard similarity between the sets of active neurons for task \textit{i} and task \textit{j} for a BNN with \textbf{(a)} Dropout, \textbf{(b)} metaplasticity, and \textbf{(c)} an engram gating module. Image is best viewed in color.}
\label{fig:fig4}
\end{figure}

We also visualize neuron overlap for each model's last hidden layer in Fig.~\ref{fig:fig4} to assess the degree of shared representations between tasks in the engramBNN, \textit{m}BNN, and BNN with Dropout. To quantify neuron reuse, we used the Jaccard overlap metric. Because the engramBNN uses explicit sparse routing, its overlap is computed directly from the top-$k$ gating masks. In contrast, the ungated baselines (\textit{m}BNN and BNN with Dropout) have zero structural sparsity. Their activity patterns are instead defined by thresholding deterministic forward activations, classifying a neuron as active only if its binary state is $+1$. Despite this baseline thresholding inherently yielding denser active populations, the engramBNN achieved higher overall accuracy and greater targeted overlap between some tasks (i.e., tasks 1 and 4) than the other architectures. In the context of preserved accuracy, this higher overlap indicates that our method facilitates constructive feature reuse, further suggesting that explicitly controlled neuron reuse through the gating module supports more robust continual learning compared to the uncontrolled activations of standard BNNs.

To evaluate the discriminative power of the learned representations, we extracted features from the final hidden layer of the Vanilla BNN, {\it m}BNN, and engramBNN following the sequential training of all Split MNIST tasks. A balanced dataset containing representative samples for all 10 classes was passed through each trained network to retrieve these representations. 

\begin{figure*}[ht!]
\centering
\includegraphics[width=1.25\textwidth]{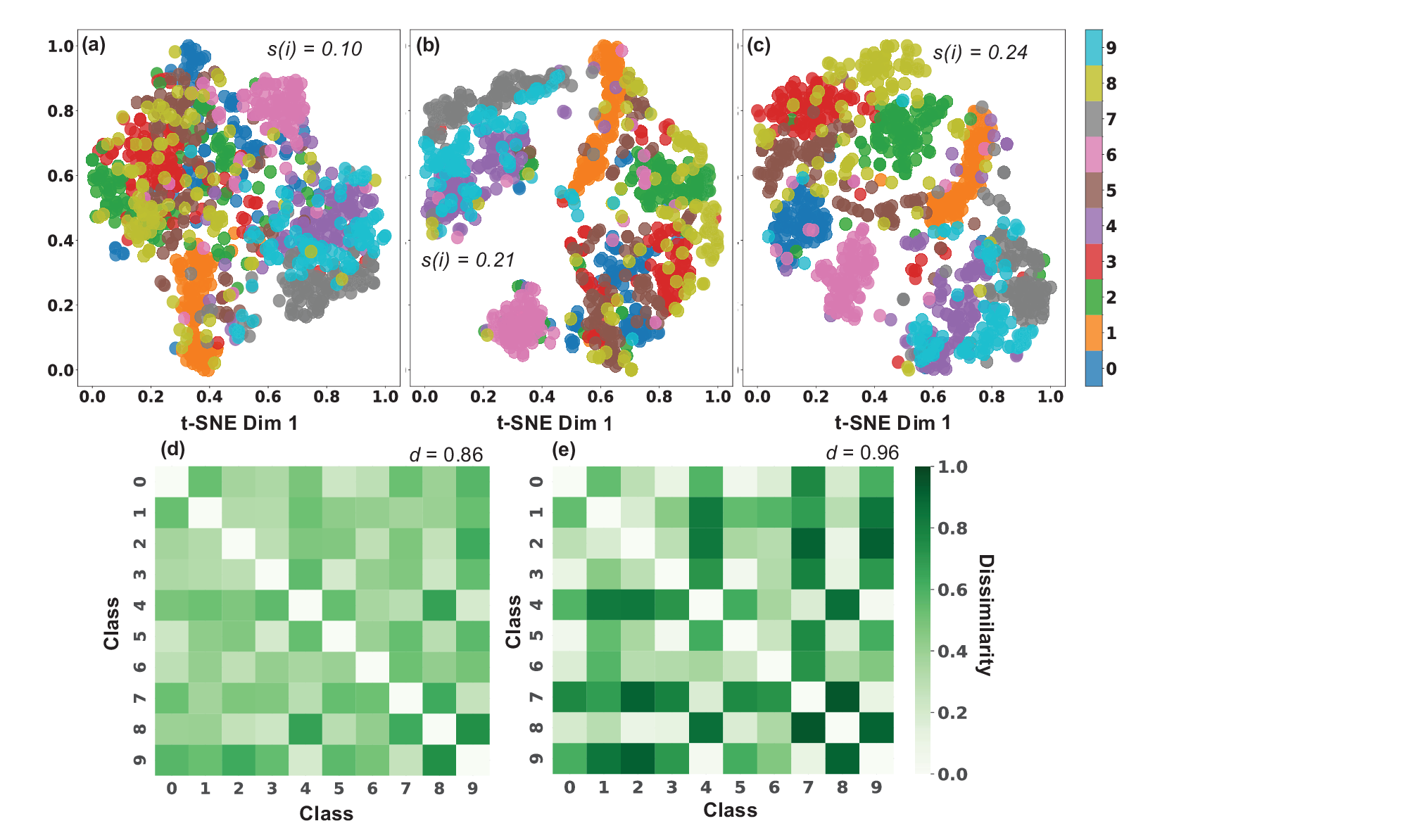}
\caption{
\textbf{Analysis of the engram block on the Split MNIST task}. After training on all sequential class-incremental Split MNIST tasks, t-SNE feature visualization and corresponding silhouette scores, $S$, were computed for extracted features from the last hidden layer of the \textbf{(a)} Vanilla BNN, \textbf{(b)} {\textit m}BNN, and \textbf{(c)} engramBNN. Each plot illustrates how samples taken from each task are clustered and separated in a 2D space, and each color is a different class. On the same task, representational dissimilarity matrix and dissimilarity score, $\bar{d}$, were visualized for the \textbf{(d)} Vanilla BNN and \textbf{(e)} engramBNN. Image is best viewed in color.}
\label{fig:fig5}
\end{figure*}

For a qualitative assessment, we applied t-SNE to project these high-dimensional embeddings into a compressed two-dimensional space, visually demonstrating how well each model separates the classes. These projections are presented in Fig.~\ref{fig:fig5}(a)-(c). The Vanilla BNN, which achieved the lowest accuracy, failed to form distinct clusters for each digit class. The {\it m}BNN (middle panel) formed tighter groupings but still displayed significant mixing between closely related classes. Among all models, engramBNN exhibited the most well-defined clusters, effectively separating previously intermixed classes that share structural similarities (e.g., digits 8, 2, 3, 5, and 0). While some overlap remains, the features are significantly more separable.

To quantify this clustering quality objectively, we calculated the Silhouette score ($S$) on the high-dimensional extracted features \cite{pedregosa2011scikit}. To assess the natural geometric separation of the latent space without relying on ground-truth supervision, we first applied $k$-means clustering \cite{hartigan1979algorithm} to the features, setting $k=10$ to correspond to the total number of digit classes. The silhouette score $s(i)$ for a single sample $i$ is defined as:

\begin{equation}
\label{eq:silhouette}
s(i) = \frac{b(i) - a(i)}{\max\{a(i), b(i)\}}
\end{equation}

\noindent
where $a(i)$ is the average distance between sample $i$ and all other points in the same cluster, and $b(i)$ is the minimum average distance between sample $i$ and all points in the closest neighboring cluster. The final score $S$ is calculated as the average of all $s(i)$ values across all samples, and is bounded between $[-1, 1]$.

A higher overall silhouette score indicates that the natural clusters are dense and well-separated. As listed in Fig.~\ref{fig:fig5}(a)-(c), engramBNN achieved the highest score of $0.24$, quantitatively confirming the superior class separation observed in the t-SNE projections, whereas the Vanilla BNN yielded the lowest score of $0.10$.

Additionally, representational dissimilarity matrices (RDMs) \cite{nili2014toolbox, greco2024predictive} are presented for the SplitMNIST experiment for the vanilla BNN and engramBNN in Fig.~\ref{fig:fig5}(d)-(e), respectively. For a set of $C$ classes, the RDM is an $C \times C$ symmetric matrix where each element quantifies the dissimilarity between the activation patterns of two classes on a scale of $0$ to $1$. Here, we use a Pearson correlation $r_{ij}$ \cite{benesty2009pearson}, to define a set of dissimilarities $d_{ij}$, which are defined in Equ.~\ref{eq:pearson} and~\ref{eq:rdm}, respectively. For all $d_{ij}$ calculations, values closer to $1$ indicate more distinct representations. To interpret overall class separation, we also computed the average off-diagonal dissimilarity $\bar{d}$, thus capturing the mean dissimilarity between all pairs of distinct classes. $\bar{d}$ was computed by averaging all $d_{i,j}$ values, excluding self-similarity.

\begin{equation}
\label{eq:pearson}
r_{ij} = \frac{\sum_{m=1}^{N} (X_{im} - \bar{X}_i)(X_{jm} - \bar{X}_j)}
               {\sqrt{\sum_{m=1}^{N} (X_{im} - \bar{X}_i)^2} \sqrt{\sum_{m=1}^{N} (X_{jm} - \bar{X}_j)^2}}
\end{equation}

\begin{equation}
\label{eq:rdm}
d_{ij} = \frac{1 - r_{ij}}{2}
\end{equation}

\noindent
Here, $X_i, X_j \in \mathbb{R}^N$ are the embeddings for each class and $\bar{X}_i$ is the mean of $X_i$, where $N$ is the dimensionality of the embeddings. $X_{im}$ is the $m$-th element of the class embedding $X_i$.

On a dissimilarity scale of $0$ to $1$, the engramBNN has a higher off-diagonal Pearson dissimilarity average, $\bar{d}$, of $0.96$ compared to the vanilla BNN score of $0.86$. These metrics reveal that the representations for each class are better separated by the engramBNN than the vanilla BNN. This closely aligns with the neural properties of the hippocampus, which enables pattern separation to prevent catastrophic interference in the brain \cite{leutgeb2007pattern}.

\subsection{System Performance and Memory Overhead}

The peak GPU and RAM usage, expressed as percentages, were also recorded during the Split MNIST experiment and are presented in Table~\ref{tab:tab5}. This was recorded using the Weights and Biases (WandB) tracking platform \protect\cite{wandb}. WandB logs system metrics using a default 10-second polling interval. Because PyTorch memory allocation reaches a stable steady state early in the training process, and individual task training durations take approximately $30$ minutes on average (totaling roughly $2.5$ hours for the $5$-task Split MNIST benchmark), this sampling frequency yields approximately $180$ steady state data points per task. 
Metrics were measured natively in PyTorch, without using a specialized low-level inference engine. This measure records the maximum total power consumption of the GPU and CPU during training for each model. Additionally, parameter count and memory size (reported in MiB) are compared with state-of-the-art models and baseline models representative of alternative continual learning approaches, namely replay-based and modular-based networks. Memory size is calculated as \( \frac{N_{\text{params}} \times b}{8 \times 10^6} \), where $N_{params}$ is the total number of model parameters, and $b$ is bit precision per parameter.
For a fair comparison, all analyzes are conducted with the class-incremental Split MNIST experiment in mind, using the same backbone architecture: two hidden layers with 2048 neurons each. Considering these metrics in the context of resource efficiency in edge devices and for real-world deployment is crucial.

For GPU usage, the engramBNN had the lowest utilization at $4.51\%$. The WSN model had the lowest RAM process usage, equal to $2.63\%$. Both models demonstrated higher efficiency compared to all other models. The model with the highest GPU usage was EWC, at $92.74\%$. The SDMLP had the highest RAM usage at $26.61\%$.

\begin{table}[t]
    \caption{Resource Utilization Metrics. Measurements were taken from the Weights and Biases (WandB \protect\cite{wandb}) system metric documentation, which is logged every 10 seconds. Metrics were measured in PyTorch, and no specialized inference engine was used. The total GPU memory is approximately 17 GB. Lowest values are bolded.}
    \label{tab:tab5}
    \begin{ruledtabular}
    \begin{tabular}{lcc}
        \textbf{Model} & \parbox[c]{3.0cm}{\centering \textbf{Peak GPU Usage (\%) $\downarrow$}} & \parbox[c]{3.0cm}{\centering \textbf{Peak RAM Usage (\%) $\downarrow$}} \\
        \hline
        Vanilla BNN & $7.95$ & $21.18$ \\
        \textbf{engramBNN} & $\mathbf{4.51}$ & $19.51$ \\
        EWC & $92.74$ & $21.79$ \\
        SI & $57.53$ & $6.66$ \\
        LwF & $12.39$ & $25.71$ \\
        XdG & $8.47$ & $22.30$ \\
        \textit{m}BNN & $8.46$ & $22.31$ \\
        SDMLP & $7.30$  & $26.61$ \\
        CH-HNN & $4.54$ & $16.98$ \\
        WSN & $21.23$ & $\mathbf{2.63}$ \\
    \end{tabular}
    \end{ruledtabular}
\end{table}

Additionally, parameter and memory metrics are presented in Table~\ref{tab:tab6} during the Split MNIST experiment. The {\it m}BNN possesses the smallest memory size of $0.69$ MiB, while the engramBNN has the second smallest memory size of $3.02$ MiB. Compared to its full-precision counterpart (that is, a model of the exact same architecture using linear layers), engramBNN has a $\sim 8\times$ smaller memory footprint.

\begin{table}[t]
    \caption{Memory overhead metrics. The number of parameters and the memory size are taken into account during inference. Memory size is recorded in MiB. Lowest values are reported in bold.}
    \label{tab:tab6}
    \begin{ruledtabular}
    \begin{tabular}{lcc}
        \textbf{Model} & \parbox[c]{2.0cm}{\centering \textbf{Parameter Count} $\downarrow$} & \parbox[c]{2.5cm}{\centering \textbf{Memory Size (MiB)$\downarrow$}} \\
        \hline
        \textbf{engramBNN} & $6,429,877$ & $3.02$ \\
        Full Precision Counterpart & $6,429,877$ & $24.53$ \\
        \textit{m}BNN & $5,820,419$ & $\mathbf{0.69}$ \\
        BI-Replay & $6,892,882$ & $26.30$ \\
        Stochastic Synapses & $17,461,248$ & $66.61$ \\
        PNN$^\ast$ & $113,003,570$ & $430.97$ \\
        SDMLP & $\mathbf{5,820,416}$  & $22.20$ \\
        CH-HNN & $5,832,714$ & $22.25$ \\
        WSN & $11,608,064$ & $44.28$ \\
    \end{tabular}
    \end{ruledtabular}
    \vspace{1ex}
    \raggedright
    {\scriptsize $^\ast$PNN metrics were calculated considering a 5-task Split MNIST experiment.}
\end{table}

The engram block adds about $600,000$ additional parameters to the {\it m}BNN, or $\sim 1.1\times$ more parameters. Furthermore, our model has $\sim 4\times$ the memory size of the {\it m}BNN. However, our model uses significantly less memory than other continuous learning models, while maintaining a comparable parameter count. Besides the engramBNN and {\it m}BNN, all other methods are estimated to take up over $20$ MiB during inference.

\section{Discussion}
\label{disc}
Metaplasticity and probabilistic context-dependent gating inform artificial engrams in a resource-efficient BNN architecture. Our model was tested in a class- and domain-incremental learning setting using image-based data. Both tests demonstrated performance comparable to the previously validated metaplastic BNN, surpassing the GPU-based efficiency of the baseline and vanilla models.  

The combination of context-dependent learning \cite{heald2023computational} and metaplasticity \cite{jedlicka2022contributions} provides a meaningful solution to continual learning by enabling more flexible and adaptive learning mechanisms. Context-dependent signals inform artificial engram modules \cite{guskjolen2023engram}. Probabilistic activations induce these masks through their stochastic nature, enabling greater knowledge retention. The probabilistic activation of engram formation facilitated stable, scalable learning, contributing to the overall effectiveness of the model \cite{ueltzhoffer2015stochastic}. The metaplasticity mechanism further adjusts synaptic plasticity based on prior learning, enabling the model to more effectively prevent task interference. Together, these processes enable more stable learning and contribute to the development of power-efficient systems with binarized architectures, making them suitable for real-world lifelong learning applications with resource constraints \cite{soltoggio2024collective}. Furthermore, our method is closely related to the recently published CH-HNN method \cite{shi2025hybrid}, which was compared in our experiments. Both approaches employ metaplasticity and induce some gating to modulate the signal. However, we argue that our approach is more efficient on the backend because it eliminates the need for pretraining and computing similarities between modulation signals. Unlike the CH-HNN, which requires hybridizing an ANN and an SNN, our method simplifies the process while achieving similar, if not improved, results.

Our model showed superior performance in the class-incremental Split MNIST experiment, outperforming vanilla models, baselines, and state-of-the-art continual learning methods, while significantly reducing resource usage, using less than half the GPU power of most other baselines and taking up only $3.02$ MiB of memory in the inference stage of Split MNIST. In edge computing and real-time applications, consideration of resource efficiency is critical \cite{shuvo2022efficient}. 

How tasks are split in class-incremental scenarios is also critical to evaluating continual learning strategies. Different splits represent varying degrees of task similarity, which may influence degrees of forgetting and knowledge transfer. In the Split CIFAR-100 experiments, neither split demonstrated superior performance over the other, suggesting that more extensive experiments with backbones forming complex architectures are warranted. However, while almost all models exhibited final accuracies near or only marginally above random chance, the engramBNN achieved the best performance, with average accuracies between $8-10\%$. This strengthens the argument for contextual gating mechanisms and is remarkable, given that a binarized, highly quantized backbone can maintain performance that fully connected networks were unable to achieve. 

In this study, we also focused on a domain-incremental setting to evaluate the model's ability to learn and retain knowledge as the data distribution changes. We strongly believe that Permuted MNIST is ineffective as a continual learning benchmark. The reason it is referred to here is its prevalent use. We employed the CORe50 dataset \cite{lomonaco2017core50} as a realistic alternative to address the need for a more realistic dataset. This benchmark posed challenges, as the images were more complex and had eleven different scenarios with varying backgrounds and lighting. Our model achieved higher weighted accuracy than other models and demonstrated forward transfer. 

EngramBNN can continually learn in the domain-incremental setting, as evident in its FWT and BWT scores. However, it struggles to capture the complexity of RGB images of diverse objects. This may be addressed by employing different backbones for our model, such as low-power CNNs, SNNs, or energy-based models \cite{li2022energy}. However, it should be acknowledged that there is a trade-off among efficiency, power consumption, and performance in meeting the low-power requirements of edge devices.

In the domain-incremental version of SplitMNIST, our model outperformed all other models, achieving a 10\% improvement over {\it m}BNN and a 6\% improvement over state-of-the-art methods SDMLP and XdG. This further validates the effectiveness of engramBNNs under domain shifts. To better maintain performance on the earliest tasks, the metaplastic $m$ hyperparameter can be increased, though this also affects peak performance on recent tasks. As with many scenarios in both continual learning and the brain, there is a balance between the stability of old information and the plasticity needed to take in new information.

The domain- and class-incremental experiments are task-agnostic because they do not require a task ID to be explicitly provided to the model (except for the XdG method, which requires it in all scenarios). Although we did not explore new strategies for real-time label generation or incorporate a noisy student approach here, we acknowledge that this is an important direction for future studies. Incorporating methods such as noisy student labeling or weak- or pseudo-labeling \cite{xie2020self} could potentially overcome this limitation, but it requires careful consideration of the trade-offs between computational efficiency and accuracy in such settings.

The semantic separation between class representations was also validated through feature visualization, quantifiable clustering, and overlap metrics. Qualitative analysis during the Split MNIST task showed that clusters were more well-established with the gating module attached, and this was further supported by a higher silhouette score than the other models. Ablation studies also revealed that our model exhibited greater Jaccard overlap in neuron masks across tasks, suggesting that the gating module encourages more relevant, shared representations while still maintaining distinct task clusters. Additionally, RDMs introduced higher overall dissimilarity in the engramBNN, which is analogous to pattern separation in the hippocampus \cite{leutgeb2007pattern}. This resembles how sparse representations and separate activations enable the brain to avoid catastrophic forgetting.

Notably, our model achieves competitive or superior performance compared to other fully connected models while consuming over $8\times$ less memory. This result demonstrates the efficiency and practicality of our architecture, especially in resource-constrained environments such as edge devices. Not only does our model significantly reduce memory overhead, but it also retains the most informative and relevant representations in a multitask environment. Reduced power consumption and faster inference times are both critical factors for on-device learning. Unlike conventional fully connected networks, which scale linearly, our model leverages both architectural sparsity and binary representations. This makes it more suitable for long-term deployment. Additionally, the fact that our model's memory advantage does not compromise performance has implications for sparser, quantized, and more efficient architectures in continual learning. It also aligns with biological insights that our brain encodes information through sparse neural codes.

Our model has demonstrated a promising approach to tackling the challenges of resource-efficient continual learning. By incorporating biologically inspired mechanisms, we achieved superior efficiency while maintaining competitive performance across class- and domain-incremental setups. While these results are encouraging, future work will address the limitations of supervised learning in these environments. In this work, we based our model on an established metaplastic binarized neural network that has been validated previously as a continual learning strategy. This provided a reliable and accepted backbone for introducing and validating the engram mechanism. However, more recent state-of-the-art binary architectures exist that have not been explored in a continual learning context. Integrating the lifelong learning mechanisms investigated here with such networks represents an important direction for further enhancing the performance and efficiency of our method. Further investigations will continue to focus on refining the gating mechanism and exploring its impact in more complex, real-world environments, such as with brain signals. This work also opens the door to scaling continuous learning systems to more complex environments or longer tasks without concern for memory cost. Since this architecture targets edge devices, future work will extend this research to more deployment-oriented settings by integrating inference engines to enable faster inference and lower energy consumption. Additionally, we plan to conduct inference benchmarking on edge devices to evaluate the proposed method's performance under realistic edge conditions.

Ultimately, this work lays the foundation for more sustainable, adaptive, efficient, and scalable models suitable for deployment in edge computing and dynamic learning environments.

\begin{acknowledgments}
The authors acknowledge the support from the Australian Research Council under Project DP230100019. I.A. acknowledges support from the Australian Government’s Research Training Program (RTP). L.F.H.C. acknowledges partial support from The University of Sydney. 
\end{acknowledgments}

\section*{Code Availability} \label{sec:codest}
The code is available on GitHub at \url{https://github.com/NeuroSyd/engramBNN}.

\section*{Data Availability} \label{sec:datast}
We employed a transparent approach, using non-exclusive, publicly accessible datasets that provide ample avenues for interested readers to reproduce results. They can be accessed via
\begin{itemize}
  \item[---] MNIST:  \url{http://yann.lecun.com/exdb/mnist}, 
  \item[---] CIFAR-100: \url{https://www.cs.toronto.edu/~kriz/cifar.html}, and 
  \item[---] CORe50:  \url{https://vlomonaco.github.io/core50}.
\end{itemize}

\section*{Conflicts of Interest} \label{sec:conflict}
The authors have no conflicts to disclose.

\appendix
\section{Permuted MNIST Experiments} \label{secA1}
Here, we present results from the domain-incremental Permuted MNIST task. Here, the SDMLP method showed higher accuracy performance and backward transfer-related scores, as shown in Table~\ref{tab:tab7}. Our engramBNN model showed slightly higher forward transfer ability than all other methods. 

We believe the SDMLP showed a greater advantage due to its use of fully connected linear layers in its architecture, as opposed to our binarized layers, which allow more precise weight updates and greater representational capacity, likely contributing to its improved performance. We remained comparable in performance to the hybrid CH-HNN model, which employs SNN and fully connected ANN architectures.

\begin{table*}[htpb!]
    \caption{Permuted MNIST continual learning metrics. All values are averages plus or minus their standard deviations.}
    \label{tab:tab7}
    \begin{ruledtabular}
    \begin{tabular}{lccccc}
        \textbf{Model} & \textbf{ACC (\%) $\uparrow$} & \textbf{\textit{w}ACC$^\ast$ (\%) $\uparrow$} & \textbf{FWT (\%) $\uparrow$} & \textbf{BWT (\%) $\uparrow$} & \textbf{REM (\%) $\uparrow$}
        \\
        \hline
        engramBNN & $52.94 \pm {\scriptstyle 0.76}$ & $36.19 \pm {\scriptstyle 1.15}$ & $\mathbf{11.27 \pm {\scriptstyle 0.84}}$ & $-45.55 \pm {\scriptstyle 20.00}$ & $54.45 \pm {\scriptstyle 20.00}$ \\
        Vanilla BNN & $45.75 \pm {\scriptstyle 0.99}$ & $24.90 \pm {\scriptstyle 1.51}$ & $10.06 \pm {\scriptstyle 1.01}$ & $-53.81 \pm {\scriptstyle 24.66}$ & $46.19 \pm {\scriptstyle 24.66}$ \\  
        LwF & $46.82 \pm {\scriptstyle 0.88}$ & $25.13 \pm {\scriptstyle1.86}$ & $10.49 \pm {\scriptstyle 0.89}$ & $-52.50 \pm {\scriptstyle 25.62}$ & $47.50 \pm {\scriptstyle 25.62}$ \\
        XdG & $56.98 \pm {\scriptstyle 1.08}$ & $40.52 \pm {\scriptstyle 2.31}$ & $10.61 \pm {\scriptstyle 1.02}$ & $-33.63 \pm {\scriptstyle 19.54}$ & $66.37 \pm {\scriptstyle 19.54}$ \\
        \textit{m}BNN & $49.62 \pm {\scriptstyle 1.04}$ & $30.22 \pm {\scriptstyle 1.52}$ & $9.92 \pm {\scriptstyle 0.99}$ & $-48.85 \pm {\scriptstyle 24.20}$ & $51.15 \pm {\scriptstyle 24.20}$ \\
        SDMLP & $\mathbf{63.27 \pm {\scriptstyle 0.91}}$ & $\mathbf{56.47 \pm {\scriptstyle 1.58}}$ & $9.58 \pm {\scriptstyle 1.13}$ & $\mathbf{-14.87 \pm {\scriptstyle 8.90}}$ &  $\mathbf{85.13 \pm {\scriptstyle 8.90}}$ \\
        CH-HNN & $50.85 \pm {\scriptstyle 1.31}$ & $35.16 \pm {\scriptstyle 3.35}$ & $9.98 \pm {\scriptstyle 0.96}$ & $-45.55 \pm {\scriptstyle 18.19}$ &  $54.45 \pm {\scriptstyle 18.19}$ \\ 
    \end{tabular}
    \end{ruledtabular}
    \vspace{1ex}
    \raggedright
    {\scriptsize \textsuperscript{a} For each model, all performance results are averaged over six training runs. The best values are in bold. $^\ast$\textbf{\textit{w}ACC}: weighted ACC.}
\end{table*}

\section{Benchmark Details} \label{secA2}

\subsection{Dataset Augmentation}
Across all experiments (Split MNIST, Split CIFAR, and CORe50), we applied standard data augmentation to the training sets prior to optimization.

\begin{itemize}
    \item \textbf{MNIST:} Each image is converted to a tensor and normalized using the standard dataset mean of $0.1307$ and standard deviation of $0.3081$. Unlike the natural image datasets, no additional spatial augmentations or color perturbations are applied.
    \item \textbf{CIFAR:} Each image is zero-padded by 4 pixels on all sides and randomly cropped back to its original $32 \times 32$ spatial resolution. Next, a random horizontal flip is applied with a probability of $0.5$. Finally, we apply a color jittering transformation that randomly perturbs the image brightness by a factor chosen uniformly from $[-63/255, +63/255]$. 
    \item \textbf{CORe50:} Each image is subjected to a random resized crop, which extracts a random patch of the original image and resizes it to $224 \times 224$ pixels (utilizing the default PyTorch scale and aspect ratio bounds). This is followed by a random horizontal flip ($p=0.5$) and a brightness perturbation identical to the CIFAR-10 setup, where the brightness factor is jittered by a factor chosen uniformly from $[-63/255, +63/255]$.
\end{itemize}

\bibliography{literature.bib}

@misc{wandb,
	author = {},
	title = {{E}xperiment {T}racking with {W}eights \& {B}iases: {T}he {A}{I} {D}eveloper {P}latform},
	howpublished = {\url{https://www.wandb.com}},
	year = {2025},
	note = {[Accessed 24-03-2025]}
}

@article{pedregosa2011scikit,
  title={Scikit-learn: {M}achine learning in {Python}},
  author={Pedregosa, Fabian and Varoquaux, Ga{\"e}l and Gramfort, Alexandre and Michel, Vincent and Thirion, Bertrand and Grisel, Olivier and Blondel, Mathieu and Prettenhofer, Peter and Weiss, Ron and Dubourg, Vincent and others},
  journal={The Journal of Machine Learning Research},
  volume={12},
  pages={2825--2830},
  year={2011}
}

@article{nili2014toolbox,
  title={A toolbox for representational similarity analysis},
  author={Nili, Hamed and Wingfield, Cai and Walther, Alexander and Su, Li and Marslen-Wilson, William and Kriegeskorte, Nikolaus},
  journal={PLoS Computational Biology},
  volume={10},
  number={4},
  year={2014},
}

@article{jeon2023distinctive,
  title={Distinctive properties of biological neural networks and recent advances in bottom-up approaches toward a better biologically plausible neural network},
  author={Jeon, Ikhwan and Kim, Taegon},
  journal={Frontiers in Computational Neuroscience},
  volume={17},
  pages={1092185},
  year={2023},
  publisher={Frontiers Media SA}
}

@article{hadsell2020embracing,
  title={Embracing change: {C}ontinual learning in deep neural networks},
  author={Hadsell, Raia and Rao, Dushyant and Rusu, Andrei A and Pascanu, Razvan},
  journal={Trends in Cognitive Sciences},
  volume={24},
  number={12},
  pages={1028--1040},
  year={2020},
  publisher={Elsevier}
}

@article{jedlicka2022contributions,
  title={Contributions by metaplasticity to solving the catastrophic forgetting problem},
  author={Jedlicka, Peter and Tomko, Matus and Robins, Anthony and Abraham, Wickliffe C},
  journal={Trends in Neurosciences},
  volume={45},
  number={9},
  pages={656--666},
  year={2022},
  publisher={Elsevier}
}

@article{wang2024comprehensive,
  title={A comprehensive survey of continual learning: {Theory}, method and application},
  author={Wang, Liyuan and Zhang, Xingxing and Su, Hang and Zhu, Jun},
  journal={IEEE Transactions on Pattern Analysis and Machine Intelligence},
  year={2024},
  volume={46},
  number={8},
  pages={5362-5383}
}

@article{kirkpatrick2017overcoming,
  title={Overcoming catastrophic forgetting in neural networks},
  author={Kirkpatrick, James and Pascanu, Razvan and Rabinowitz, Neil and Veness, Joel and Desjardins, Guillaume and Rusu, Andrei A and Milan, Kieran and Quan, John and Ramalho, Tiago and Grabska-Barwinska, Agnieszka and others},
  journal={Proceedings of the National Academy of Sciences},
  volume={114},
  number={13},
  pages={3521--3526},
  year={2017}
}

@article{zenke2017continual,
  title={Continual learning through synaptic intelligence},
  author={Zenke, Friedemann and Poole, Ben and Ganguli, Surya},
  journal={International Conference on Machine Learning},
  pages={3987--3995},
  year={2017},
  organization={PMLR}
}

@article{li2017learning,
  title={Learning without forgetting},
  author={Li, Zhizhong and Hoiem, Derek},
  journal={IEEE Transactions on Pattern Analysis and Machine Intelligence},
  volume={40},
  number={12},
  pages={2935--2947},
  year={2017}
}

@article{van2020brain,
  title={Brain-inspired replay for continual learning with artificial neural networks},
  author={Van de Ven, Gido M and Siegelmann, Hava T and Tolias, Andreas S},
  journal={Nature Communications},
  volume={11},
  number={1},
  pages={4069},
  year={2020}
}

@article{rebuffi2017icarl,
  title={i{C}a{RL}: {I}ncremental classifier and representation learning},
  author={Rebuffi, Sylvestre-Alvise and Kolesnikov, Alexander and Sperl, Georg and Lampert, Christoph H},
  journal={Proceedings of the IEEE Conference on Computer Vision and Pattern Recognition},
  pages={2001--2010},
  year={2017}
}

@article{lopez2017gradient,
  title={Gradient episodic memory for continual learning},
  author={Lopez-Paz, David and Ranzato, Marc'Aurelio},
  journal={Advances in Neural Information Processing Systems},
  volume={30},
  year={2017}
}

@article{mallya2018piggyback,
  title={Piggyback: {A}dapting a single network to multiple tasks by learning to mask weights},
  author={Mallya, Arun and Davis, Dillon and Lazebnik, Svetlana},
  journal={Proceedings of the European Conference on Computer Vision (ECCV)},
  pages={67--82},
  year={2018}
}

@article{mallya2018packnet,
  title={Packnet: {A}dding multiple tasks to a single network by iterative pruning},
  author={Mallya, Arun and Lazebnik, Svetlana},
  journal={Proceedings of the IEEE Conference on Computer Vision and Pattern Recognition},
  pages={7765--7773},
  year={2018}
}

@article{zeng2019continual,
  title={Continual learning of context-dependent processing in neural networks},
  author={Zeng, Guanxiong and Chen, Yang and Cui, Bo and Yu, Shan},
  journal={Nature Machine Intelligence},
  volume={1},
  number={8},
  pages={364--372},
  year={2019}
}

@article{javed2019meta,
  title={Meta-learning representations for continual learning},
  author={Javed, Khurram and White, Martha},
  journal={Advances in Neural Information Processing Systems},
  volume={32},
  pages={1818–1828},
  year={2019}
}

@article{madaan2021representational,
  title={Representational continuity for unsupervised continual learning},
  author={Madaan, Divyam and Yoon, Jaehong and Li, Yuanchun and Liu, Yunxin and Hwang, Sung Ju},
  journal={preprint arXiv:2110.06976},
  year={2021}
}

@article{abraham1996metaplasticity,
  title={Metaplasticity: {T}he plasticity of synaptic plasticity},
  author={Abraham, Wickliffe C and Bear, Mark F},
  journal={Trends in Neurosciences},
  volume={19},
  number={4},
  pages={126--130},
  year={1996}
}

@article{abraham2008metaplasticity,
  title={Metaplasticity: {T}uning synapses and networks for plasticity},
  author={Abraham, Wickliffe C},
  journal={Nature Reviews Neuroscience},
  volume={9},
  number={5},
  pages={387--387},
  year={2008}
}

@article{guskjolen2023engram,
  title={Engram neurons: {E}ncoding, consolidation, retrieval, and forgetting of memory},
  author={Guskjolen, Axel and Cembrowski, Mark S},
  journal={Molecular Psychiatry},
  volume={28},
  number={8},
  pages={3207--3219},
  year={2023}
}

@article{reijmers2007localization,
  title={Localization of a stable neural correlate of associative memory},
  author={Reijmers, Leon G and Perkins, Brian L and Matsuo, Naoki and Mayford, Mark},
  journal={Science},
  volume={317},
  number={5842},
  pages={1230--1233},
  year={2007}
}

@article{tayler2013reactivation,
  title={Reactivation of neural ensembles during the retrieval of recent and remote memory},
  author={Tayler, Kaycie K and Tanaka, Kazumasa Z and Reijmers, Leon G and Wiltgen, Brian J},
  journal={Current Biology},
  volume={23},
  number={2},
  pages={99--106},
  year={2013}
}

@article{wasserman1997whats,
  title={What's elementary about associative learning?},
  author={Wasserman, Edward A and Miller, Ralph R},
  journal={Annual Review of Psychology},
  volume={48},
  number={1},
  pages={573--607},
  year={1997}
}

@article{thompson1997associative,
  title={Associative learning},
  author={Thompson, Richard F and Bao, Shaowen and Chen, Lu and Cipriano, Benjamin D and Grethe, Jeffrey S and Kim, Jeansok J and Thompson, Judith K and Tracy, Jo Anne and Weninger, Martha S and Krupa, David J},
  journal={International Review of Neurobiology},
  volume={41},
  pages={151--189},
  year={1997}
}

@article{heald2023computational,
  title={The computational and neural bases of context-dependent learning},
  author={Heald, James B and Wolpert, Daniel M and Lengyel, M{\'a}t{\'e}},
  journal={Annual Review of Neuroscience},
  volume={46},
  number={1},
  pages={233--258},
  year={2023}
}

@article{gershman2015unifying,
  title={A unifying probabilistic view of associative learning},
  author={Gershman, Samuel J},
  journal={PLoS Computational Biology},
  volume={11},
  number={11},
  pages={e1004567},
  year={2015}
}

@article{liljenstrom1995noise,
  title={Noise-enhanced performance in a cortical associative memory model},
  author={Liljenstr{\"o}m, Hans and Wu, X-B},
  journal={International Journal of Neural Systems},
  volume={6},
  number={01},
  pages={19--29},
  year={1995}
}

@article{dong2020survey,
  title={A survey on ensemble learning},
  author={Dong, Xibin and Yu, Zhiwen and Cao, Wenming and Shi, Yifan and Ma, Qianli},
  journal={Frontiers of Computer Science},
  volume={14},
  pages={241--258},
  year={2020}
}

@article{srivastava2014dropout,
  title={Dropout: {A} simple way to prevent neural networks from overfitting},
  author={Srivastava, Nitish and Hinton, Geoffrey and Krizhevsky, Alex and Sutskever, Ilya and Salakhutdinov, Ruslan},
  journal={The Journal of Machine Learning Research},
  volume={15},
  number={1},
  pages={1929--1958},
  year={2014}
}

@article{louizos2017learning,
  title={Learning sparse neural networks through $L_0$ regularization},
  author={Louizos, Christos and Welling, Max and Kingma, Diederik P},
  journal={The Fifth International Conference on Learning Representations},
  year={2017},
  organization={ICLR}
}

@article{pan2009survey,
  title={A survey on transfer learning},
  author={Pan, Sinno Jialin and Yang, Qiang},
  journal={IEEE Transactions on Knowledge and Data Engineering},
  volume={22},
  number={10},
  pages={1345--1359},
  year={2009}
}

@article{rizzuto2001autoassociative,
  title={An autoassociative neural network model of paired-associate learning},
  author={Rizzuto, Daniel S and Kahana, Michael J},
  journal={Neural Computation},
  volume={13},
  number={9},
  pages={2075--2092},
  year={2001}
}

@article{tang2013learning,
  title={Learning stochastic feedforward neural networks},
  author={Tang, Charlie and Salakhutdinov, Russ R},
  journal={Advances in Neural Information Processing Systems},
  volume={26},
  year={2013}
}

@article{masse2018alleviating,
  title={Alleviating catastrophic forgetting using context-dependent gating and synaptic stabilization},
  author={Masse, Nicolas Y and Grant, Gregory D and Freedman, David J},
  journal={Proceedings of the National Academy of Sciences},
  volume={115},
  number={44},
  pages={e10467--e10475},
  year={2018},

}

@article{tilley2023artificial,
    title={Artificial Neuronal Ensembles with Learned Context Dependent Gating},
    author={Matthew James Tilley and Michelle Miller and David Freedman},
    journal={The Eleventh International Conference on Learning Representations},
    year={2023},
    organization={ICLR}
}

@article{courbariaux2015binaryconnect,
author = {Courbariaux, Matthieu and Bengio, Yoshua and David, Jean-Pierre},
title = {Binary{C}onnect: {T}raining deep neural networks with binary weights during propagations},
year = {2015},
publisher = {MIT Press},
address = {Cambridge, MA, USA},
booktitle = {Proceedings of the 29th International Conference on Neural Information Processing Systems},
pages = {3123–3131},
volume={2},
numpages = {9},
location = {Montreal, Canada},
}

@article{conti2018xnor,
  title={{XNOR} neural engine: {A} hardware accelerator {IP} for 21.6-f{J}/op binary neural network inference},
  author={Conti, Francesco and Schiavone, Pasquale Davide and Benini, Luca},
  journal={IEEE Transactions on Computer-Aided Design of Integrated Circuits and Systems},
  volume={37},
  number={11},
  pages={2940--2951},
  year={2018}
}

@article{hubara2016binarized,
  title={Binarized neural networks},
  author={Hubara, Itay and Courbariaux, Matthieu and Soudry, Daniel and El-Yaniv, Ran and Bengio, Yoshua},
  journal={Advances in Neural Information Processing Systems},
  volume={29},
  pages={4107–4115},
  year={2016}
}

@article{simons2019review,
  title={A Review of Binarized Neural Networks},
  author={Simons, Taylor and Lee, Dah-Jye},
  journal={Electronics},
  volume={8},
  number={6},
  pages={661},
  year={2019}
}

@article{rastegari2016xnor,
  title={{XNOR-Net}: {Imagenet} classification using binary convolutional neural networks},
  author={Rastegari, Mohammad and Ordonez, Vicente and Redmon, Joseph and Farhadi, Ali},
  journal={European Conference on Computer Vision},
  pages={525--542},
  year={2016}
}

@article{laborieux2021synaptic,
  title={Synaptic metaplasticity in binarized neural networks},
  author={Laborieux, Axel and Ernoult, Maxence and Hirtzlin, Tifenn and Querlioz, Damien},
  journal={Nature Communications},
  volume={12},
  number={1},
  pages={2549},
  year={2021}
}

@article{aguilar2025continuous,
  title={Continuous metaplastic training on brain signals},
  author={Aguilar, Isabelle and Bersani--Veroni, Thomas and Herbozo Contreras, Luis Fernando and Nikpour, Armin and Querlioz, Damien and Kavehei, Omid},
  journal={npj Unconventional Computing},
  volume={2},
  pages={1-9},
  year={2025}
}

@article{xie2020self,
  title={Self-training with noisy student improves {ImageNet} classification},
  author={Xie, Qizhe and Luong, Minh-Thang and Hovy, Eduard and Le, Quoc V},
  journal={Proceedings of the IEEE/CVF Conference on Computer Vision and Pattern Recognition},
  pages={10687--10698},
  year={2020}
}

@article{farquhar2018towards,
  title={Towards robust evaluations of continual learning},
  author={Farquhar, Sebastian and Gal, Yarin},
  journal={Lifelong Learning: A Reinforcement Learning Approach Workshop, International Conference on Machine Learning},
  year={2018},
  organization={PMLR}
}

@article{lomonaco2017core50,
  title={{CORe50}: {A} new dataset and benchmark for continuous object recognition},
  author={Lomonaco, Vincenzo and Maltoni, Davide},
  journal={Conference on Robot Learning},
  pages={17--26},
  year={2017}
}

@article{li2022energy,
  title={Energy-based models for continual learning},
  author={Li, Shuang and Du, Yilun and Van de Ven, Gido and Mordatch, Igor},
  journal={Conference on Lifelong Learning Agents},
  pages={1-22},
  year={2022}
}

@inproceedings{basso2024class,
  title={On Class-Incremental Learning for Fully Binarized Convolutional Neural Networks},
  author={Basso-Bert, Yanis and Guicquero, William and Molnos, Anca and Lemaire, Romain and Dupret, Antoine},
  booktitle={IEEE International Symposium on Circuits and Systems (ISCAS)},
  pages={1--5},
  year={2024},
}

@article{vorabbi2024enabling,
  title={Enabling on-device continual learning with binary neural networks and latent replay},
  author={Vorabbi, Lorenzo and Maltoni, Davide and Borghi, Guido and Santi, Stefano and others},
  booktitle={Proceedings of the 19th International Joint Conference on Computer Vision, Imaging and Computer Graphics Theory and Applications - VISAPP},
  volume={2},
  pages={25--36},
  year={2024},
  organization={Science and Technology Publications}
}

@article{shi2025hybrid,
  title={Hybrid neural networks for continual learning inspired by corticohippocampal circuits},
  author={Shi, Qianqian and Liu, Faqiang and Li, Hongyi and Li, Guangyu and Shi, Luping and Zhao, Rong},
  journal={Nature Communications},
  volume={16},
  number={1},
  pages={1272},
  year={2025},
  publisher={Nature Publishing Group UK London}
}

@inproceedings{bricken2023sparse,
  title={Sparse Distributed Memory is a Continual Learner},
  author={Bricken, Trenton and Davies, Xander and Singh, Deepak and Krotov, Dmitry and Kreiman, Gabriel},
  booktitle={International Conference on Learning Representations},
  year={2023},
  organization={ICLR}
}

@inproceedings{kang2022forget,
  title={Forget-free continual learning with winning subnetworks},
  author={Kang, Haeyong and Mina, Rusty John Lloyd and Madjid, Sultan Rizky Hikmawan and Yoon, Jaehong and Hasegawa-Johnson, Mark and Hwang, Sung Ju and Yoo, Chang D},
  booktitle={International Conference on Machine Learning},
  pages={10734--10750},
  year={2022},
  organization={PMLR}
}

@article{ye2023continual,
  title={Continual variational autoencoder via continual generative knowledge distillation},
  author={Ye, Fei and Bors, Adrian G},
  booktitle={Proceedings of the AAAI Conference on Artificial Intelligence},
  volume={37},
  pages={10918--10926},
  year={2023}
}

@inproceedings{deja2021binplay,
  title={Binplay: A binary latent autoencoder for generative replay continual learning},
  author={Deja, Kamil and Wawrzy{\'n}ski, Pawe{\l} and Marczak, Daniel and Masarczyk, Wojciech and Trzci{\'n}ski, Tomasz},
  booktitle={IEEE International Joint Conference on Neural Networks (IJCNN)},
  pages={1--8},
  year={2021}
}

@article{shi2023unified,
  title={A unified approach to domain incremental learning with memory: {T}heory and algorithm},
  author={Shi, Haizhou and Wang, Hao},
  journal={Advances in Neural Information Processing Systems},
  volume={36},
  pages={15027--15059},
  year={2023}
}

@inproceedings{zohora2021metaplasticnet,
  title={Metaplastic{N}et: {A}rchitecture with probabilistic metaplastic synapses for continual learning},
  author={Zohora, Fatima Tuz and Karia, Vedant and Daram, Anurag Reddy and Zyarah, Abdullah M and Kudithipudi, Dhireesha},
  booktitle={IEEE International Symposium on Circuits and Systems (ISCAS)},
  pages={1--5},
  year={2021}
}

@article{cho2025neuromimetic,
  title={Neuromimetic metaplasticity for adaptive continual learning without catastrophic forgetting},
  author={Cho, Suhee and Lee, Hyeonsu and Baek, Seungdae and Paik, Se-Bum},
  journal={Neural Networks},
  pages={107762},
  year={2025},
  publisher={Elsevier}
}

@article{rusu2016progressive,
  title={Progressive neural networks},
  author={Rusu, Andrei A and Rabinowitz, Neil C and Desjardins, Guillaume and Soyer, Hubert and Kirkpatrick, James and Kavukcuoglu, Koray and Pascanu, Razvan and Hadsell, Raia},
  journal={preprint arXiv:1606.04671},
  year={2016}
}

@article{schug2021presynaptic,
  title={Presynaptic stochasticity improves energy efficiency and helps alleviate the stability-plasticity dilemma},
  author={Schug, Simon and Benzing, Frederik and Steger, Angelika},
  journal={eLife},
  volume={10},
  pages={e69884},
  year={2021},
  publisher={eLife Sciences Publications Limited}
}

@article{lauriola2022introduction,
  title={An introduction to deep learning in natural language processing: {M}odels, techniques, and tools},
  author={Lauriola, Ivano and Lavelli, Alberto and Aiolli, Fabio},
  journal={Neurocomputing},
  volume={470},
  pages={443--456},
  year={2022},
  publisher={Elsevier}
}

@article{bachute2021autonomous,
  title={Autonomous driving architectures: {I}nsights of machine learning and deep learning algorithms},
  author={Bachute, Mrinal R and Subhedar, Javed M},
  journal={Machine Learning with Applications},
  volume={6},
  pages={100164},
  year={2021},
  publisher={Elsevier}
}

@inproceedings{shinde2018review,
  title={A review of machine learning and deep learning applications},
  author={Shinde, Pramila P and Shah, Seema},
  booktitle={IEEE Fourth International Conference on Computing Communication Control and Automation (ICCUBEA)},
  pages={1--6},
  year = {2018}
}

@article{mccloskey1989catastrophic,
  title={Catastrophic interference in connectionist networks: {T}he sequential learning problem},
  author={McCloskey, Michael and Cohen, Neal J},
  booktitle={Psychology of Learning and Motivation},
  volume={24},
  pages={109--165},
  year={1989},
  publisher={Elsevier}
}

@article{lee2019overview,
  title={An overview of data telemetry in inductively powered implantable biomedical devices},
  author={Lee, Byunghun and Ghovanloo, Maysam},
  journal={IEEE Communications Magazine},
  volume={57},
  number={2},
  pages={74--80},
  year={2019}
}

@inproceedings{salam2016tradeoffs,
  title={Tradeoffs between wireless communication and computation in closed-loop implantable devices},
  author={Salam, M Tariqus and Kassiri, Hossein and Soltani, Nima and He, Haoyu and Velazquez, Jose Luis Perez and Genov, Roman},
  booktitle={IEEE International Symposium on Circuits and Systems (ISCAS)},
  pages={1838--1841},
  year={2016}
}

@article{mcgann2008adaptive,
  title={Adaptive control for autonomous underwater vehicles},
  author={McGann, Conor and Py, Frederic and Rajan, Kanna and Ryan, John P and Henthorn, Richard},
  booktitle={Proceedings of the AAAI Conference on Artificial Intelligence},
  pages={1319--1324},
  year={2008}
}

@article{lee2020clinical,
  title={Clinical applications of continual learning machine learning},
  author={Lee, Cecilia S and Lee, Aaron Y},
  journal={The Lancet Digital Health},
  volume={2},
  number={6},
  pages={e279--e281},
  year={2020},
  publisher={Elsevier}
}

@book{o2012computational,
  author    = {O'Reilly, Randall C. and Munakata, Yuko and Frank, Michael J. and Hazy, Thomas E.},
  title     = {Computational Cognitive Neuroscience},
  year      = {2012},
  publisher = {PediaPress},
  address   = {Mainz, Germany},
  note      = {Version 1.0}
}

@article{rao2021synaptic,
  title={A synaptic framework for the persistence of memory engrams},
  author={Rao-Ruiz, Priyanka and Visser, Esther and Mitri{\'c}, Miodrag and Smit, August B and Van den Oever, Michel C},
  journal={Frontiers in Synaptic Neuroscience},
  volume={13},
  pages={661476},
  year={2021},
  publisher={Frontiers Media SA}
}

@article{autore2023adaptive,
  title={Adaptive expression of engrams by retroactive interference},
  author={Autore, Livia and O’Leary, James D and Ortega-de San Luis, Clara and Ryan, Tom{\'a}s J},
  journal={Cell Reports},
  volume={42},
  number={8},
  year={2023},
  publisher={Elsevier}
}

@article{fusi2005cascade,
  title={Cascade models of synaptically stored memories},
  author={Fusi, Stefano and Drew, Patrick J and Abbott, Larry F},
  journal={Neuron},
  volume={45},
  number={4},
  pages={599--611},
  year={2005},
  publisher={Elsevier}
}

@article{soltoggio2024collective,
  title={A collective {AI} via lifelong learning and sharing at the edge},
  author={Soltoggio, Andrea and Ben-Iwhiwhu, Eseoghene and Braverman, Vladimir and Eaton, Eric and Epstein, Benjamin and Ge, Yunhao and Halperin, Lucy and How, Jonathan and Itti, Laurent and Jacobs, Michael A and others},
  journal={Nature Machine Intelligence},
  volume={6},
  number={3},
  pages={251--264},
  year={2024},
  publisher={Nature Publishing Group UK London}
}

@article{ueltzhoffer2015stochastic,
  title={Stochastic dynamics underlying cognitive stability and flexibility},
  author={Ueltzh{\"o}ffer, Kai and Armbruster-Gen{\c{c}}, Diana JN and Fiebach, Christian J},
  journal={PLoS Computational Biology},
  volume={11},
  number={6},
  pages={e1004331},
  year={2015},
  publisher={Public Library of Science San Francisco, CA USA}
}

@article{shuvo2022efficient,
  title={Efficient acceleration of deep learning inference on resource-constrained edge devices: {A} review},
  author={Shuvo, Md Maruf Hossain and Islam, Syed Kamrul and Cheng, Jianlin and Morshed, Bashir I},
  journal={Proceedings of the IEEE},
  volume={111},
  number={1},
  pages={42--91},
  year={2022},
}

@article{hartigan1979algorithm,
  title={Algorithm {AS} 136: {A} k-means clustering algorithm},
  author={Hartigan, John A and Wong, Manchek A},
  journal={Journal of the Royal Statistical Society. Series C (Applied Statistics)},
  volume={28},
  number={1},
  pages={100--108},
  year={1979},
  publisher={JSTOR}
}

@incollection{benesty2009pearson,
  title={Pearson correlation coefficient},
  author={Benesty, Jacob and Chen, Jingdong and Huang, Yiteng and Cohen, Israel},
  booktitle={Noise Reduction in Speech Processing},
  pages={1--4},
  year={2009},
  publisher={Springer}
}

@article{greco2024predictive,
  title={Predictive learning shapes the representational geometry of the human brain},
  author={Greco, Antonino and Moser, Julia and Preissl, Hubert and Siegel, Markus},
  journal={Nature Communications},
  volume={15},
  number={1},
  pages={9670},
  year={2024},
  publisher={Nature Publishing Group UK London}
}

@article{leutgeb2007pattern,
  title={Pattern separation in the dentate gyrus and {CA3} of the hippocampus},
  author={Leutgeb, Jill K and Leutgeb, Stefan and Moser, May-Britt and Moser, Edvard I},
  journal={Science},
  volume={315},
  number={5814},
  pages={961--966},
  year={2007},
  publisher={American Association for the Advancement of Science}
}

@article{josselyn2015finding,
  title={Finding the engram},
  author={Josselyn, Sheena A and K{\"o}hler, Stefan and Frankland, Paul W},
  journal={Nature Reviews Neuroscience},
  volume={16},
  number={9},
  pages={521--534},
  year={2015},
  publisher={Nature Publishing Group UK London}
}

@article{rolnick2019experience,
  title={Experience replay for continual learning},
  author={Rolnick, David and Ahuja, Arun and Schwarz, Jonathan and Lillicrap, Timothy and Wayne, Gregory},
  journal={Advances in Neural Information Processing Systems},
  volume={32},
  year={2019}
}

@article{liu2012optogenetic,
  title={Optogenetic stimulation of a hippocampal engram activates fear memory recall},
  author={Liu, Xu and Ramirez, Steve and Pang, Petti T and Puryear, Corey B and Govindarajan, Arvind and Deisseroth, Karl and Tonegawa, Susumu},
  journal={Nature},
  volume={484},
  number={7394},
  pages={381--385},
  year={2012},
  publisher={Nature Publishing Group UK London}
}

\end{document}